\documentclass[conference]{IEEEtran}
\usepackage{times}

\usepackage[numbers]{natbib}
\usepackage{multicol}
\usepackage{multirow}
\usepackage[bookmarks=true]{hyperref}
\usepackage{amssymb}
\usepackage{booktabs}
\usepackage{tabularx}
\usepackage{pifont}  % for checkmarks
\usepackage{graphicx}
\usepackage{float}  % for floating figure
\usepackage{lipsum}
\usepackage{xcolor}
\let\labelindent\relax
\usepackage{enumitem}
\usepackage{siunitx}
\usepackage{algorithm}
\usepackage{algpseudocode}
\usepackage{color, colortbl}

\newcommand{\rev}[1]{\textcolor{black}{#1}}

\definecolor{LightCyan}{rgb}{0.8,1,1}

\usepackage{mathtools}

\DeclareMathOperator*{\bD}{\mathcal{D}}

\DeclareMathOperator*{\priorD}{\bD^{s}}
\DeclareMathOperator*{\targetD}{\bD^{t}}
\DeclareMathOperator*{\targetDtargetT}{\bD^{t}_{target}}\DeclareMathOperator*{\targetDpriorT}{\bD^{t}_{prior}}

\begin{document}

% \title{Using Language to Learn Domain-Invariant Representations for Sim2Real Transfer}
\title{Natural Language Can Help Bridge \\ the Sim2Real Gap}

% You will get a Paper-ID when submitting a pdf file to the conference system
% \author{Author Names Omitted for Anonymous Review. Paper-ID 164}
\author{Author Names Omitted for Anonymous Review.}

% \author{\authorblockN{Michael Shell}
% \authorblockA{School of Electrical and\\Computer Engineering\\
% Georgia Institute of Technology\\
% Atlanta, Georgia 30332--0250\\
% Email: mshell@ece.gatech.edu}
% \and
% \authorblockN{Homer Simpson}
% \authorblockA{Twentieth Century Fox\\
% Email: homer@thesimpsons.com}
% \and
% \authorblockN{James Kirk\\ and Montgomery Scott}
% \authorblockA{Starfleet Academy\\
% San Francisco, California 96678-2391\\

% avoiding spaces at the end of the author lines is not a problem with
% conference papers because we don't use \thanks or \IEEEmembership

% \author{\authorblockN{Michael Shell\authorrefmark{1},
% Homer Simpson\authorrefmark{2},
% James Kirk\authorrefmark{3}, 
% Montgomery Scott\authorrefmark{3} and
% Eldon Tyrell\authorrefmark{4}}
% \authorblockA{\authorrefmark{1}School of Electrical and Computer Engineering\\
% Georgia Institute of Technology,
% Atlanta, Georgia 30332--0250\\ Email: mshell@ece.gatech.edu}
% \authorblockA{\authorrefmark{2}Twentieth Century Fox, Springfield, USA\\
% Email: homer@thesimpsons.com}
% \authorblockA{\authorrefmark{3}Starfleet Academy, San Francisco, California 96678-2391\\
% Telephone: (800) 555--1212, Fax: (888) 555--1212}
% \authorblockA{\authorrefmark{4}Tyrell Inc., 123 Replicant Street, Los Angeles, California 90210--4321}}

\author{\authorblockN{Albert Yu,
Adeline Foote,
Raymond Mooney, and
Roberto Mart\'in-Mart\'in}
\authorblockA{UT Austin\\
\texttt{\{albertyu, addiefoote, mooney\}@utexas.edu, robertomm@cs.utexas.edu}}
}

\maketitle

\begin{abstract}
The main challenge in learning image-conditioned robotic policies is acquiring a visual representation conducive to low-level control. Due to the high dimensionality of the image space, learning a good visual representation requires a considerable amount of visual data. However, when learning in the real world, data is expensive. Sim2Real is a promising paradigm for overcoming data scarcity in the real-world target domain by using a simulator to collect large amounts of cheap data closely related to the target task. However, it is difficult to transfer an image-conditioned policy from sim to real when the domains are very visually dissimilar. To bridge the sim2real visual gap, we propose using natural language descriptions of images as a unifying signal across domains that captures the underlying task-relevant semantics. Our key insight is that if two image observations from different domains are labeled with similar language, the policy should predict similar action distributions for both images. We demonstrate that training the image encoder to predict the language description or the distance between descriptions of a sim or real image serves as a useful, data-efficient pretraining step that helps learn a domain-invariant image representation. We can then use this image encoder as the backbone of an IL policy trained simultaneously on a large amount of simulated and a handful of real demonstrations. Our approach outperforms widely used prior sim2real methods and strong vision-language pretraining baselines like CLIP and R3M by 25 to 40\%.
See additional videos and materials at \url{https://robin-lab.cs.utexas.edu/lang4sim2real/}.
\end{abstract}

\IEEEpeerreviewmaketitle
\newcommand\ouralgo{Lang4Sim2Real}

\section{Introduction}
\label{sec:intro}
Recently, visual imitation learning (IL) has achieved significant success on manipulation tasks in household environments~\cite{schaal1999imitation, brohan2022rt1}.
% \roberto{add more. we don't want to piss off people}
However, these methods rely on large amounts of data in very similar domains to train data-hungry image-conditioned policies~\citep{brohan2022rt1, brohan2023rt2, padalkar2023rtx}.
% While collecting a large number of images in the target domain (e.g., a specific home) may be feasible, collecting many image-action demonstrations in this target domain is prohibitively expensive.
Some researchers are attempting to generalize visual IL to any target domain by collecting large datasets of demonstrations from mixed domains. In this work, we explore a different approach: can we transfer a policy trained on cheaply acquired, diverse simulation data to a real-world target task with just a few demonstrations?

% This often restricts us to the few-shot imitation-learning (IL) problem setup. However, the limited number of demonstrations we can acquire is not enough to train data-hungry image-conditioned policies to perform useful tasks, such as cooking eggs.

% In such settings, it is often much cheaper to collect a large amount of simulated data on tasks similar to cooking eggs. Thus, we approach the few-shot real-world IL problem as a sim2real problem. However, s
% A sim2real approach, however, poses new difficulties. After training on lots of simulated demonstrations, adapting the network to a small number of real-world demonstrations involves overcoming a significant distribution shift caused by the large visual dissimilarity between the simulated and real domains.
A solution to effectively leverage cheap sim data while successfully fitting scarce real-world demonstrations is to create a domain-agnostic visual representation and use it for policy training. Such a representation should enable the policy to use the simulation image-action data as an inductive bias to learn with few-shot real world data. This representation must allow the policy to tap into the right distribution of actions by being broad enough to capture the task-relevant semantic state from image observations, yet fine-grained enough to be conducive to low-level control. For instance, a sim and real image observation, both showing the robot gripper a few inches above a pan handle, should lie close together in the image embedding space to lead to similar actions, even if the two images have large differences in pixel space.

\begin{figure}[t]
    \centering
    \includegraphics[width=1\linewidth]{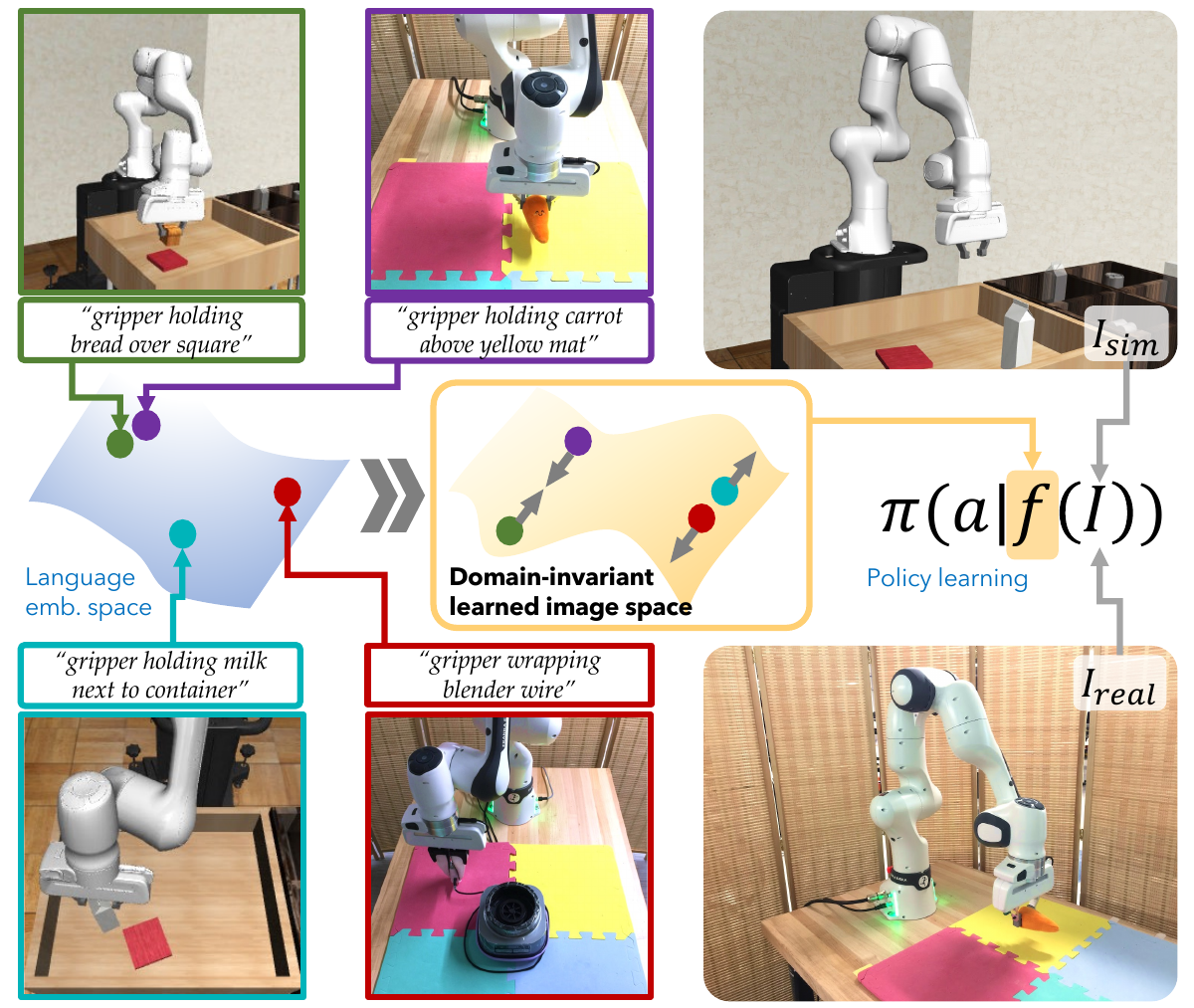}
    \caption{\textbf{Bridging the sim2real gap with language}. Robot images from simulation and the real world with similar language descriptions (\textit{green \& purple borders}) are mapped to similar features in language embedding space, while sim and real images with different language descriptions (\textit{teal \& red}) are mapped to faraway locations. We propose using language embedding similarities to re-shape the image embeddings (\textit{center}) to create a domain-invariant image space. A policy is learned conditioned on these image embeddings from both sim and real images (\textit{right}).}
    \label{fig:pullfig} 
\end{figure}
% we see two (image, language description) pairs: one from sim and one from real. These pairs show the robot at a similar position in relation to the objects, and hence the language descriptions of the two images are similar, as measured by distance in LLM space (blue). Hence we want their associated image representations to also be encoded in a similar region of the learned domain-invariant visual representation space (gold). Conversely, on the \textbf{right}, if a sim and real image observation shows the robot at a very different spatial position in relation to the objects on the scene, their language descriptions would be far apart in LLM space, and our approach pulls away the image embeddings of the corresponding image observations.

How might we acquire supervision for learning such a visual representation? Language is an ideal medium for providing it. Descriptions of task-relevant features in image observations, such as whether or not a gripper is close to a pan handle, serve as a unifying signal to align the representations of images between sim and real. We hypothesize that if a sim and real image have similar language descriptions (e.g., ``the gripper is open and right above the pan handle''), then their underlying semantic states are also similar, and thus the actions the policy predicts conditioned on each image should also be semantically similar (e.g., moving downward to reach the pan handle). The pretrained embedding space of large language models (LLMs) offers a well-tuned signal that can be leveraged to measure the semantic similarity between real and sim images via their associated language descriptions (see Fig.~\ref{fig:pullfig}). This simple insight allows us to learn a domain-agnostic visual representation to bridge the visual sim2real gap.

A popular paradigm in foundation model research is to first pretrain the backbone on large datasets, and then add and train a task-specific head to process the backbone outputs to perform a downstream task. We borrow from this paradigm by first pretraining an image encoder to predict the pretrained embeddings of language descriptions of images from roughly a few hundred trajectories in sim and real, with language labels on each image. Then we use this image encoder as the backbone of our IL policy and train on action-labeled data from both the sim and real domains simultaneously, where we only need a few action-labeled demonstrations from the real world.

In this paper, we introduce \ouralgo, a lightweight framework for transferring between any two domains that have large visual differences but contain data across a similar distribution of tasks. Our approach has the following main advantages over prior sim2real efforts:
\begin{enumerate}[leftmargin=*]
    \item Alleviates the need for the engineering-intensive task of system identification, or more broadly trying to exactly match a sim environment to the real environment both visually and semantically.
    \item Enables sim2real transfer on tasks involving deformable objects that are hard to simulate with the same dynamics and visual appearance as the real-world version of the objects.
    \item Bridges a wide sim2real gap that includes differences in: camera point-of-view (1st vs 3rd person), friction and damping coefficients, task goals, robot control frequencies, and initial robot and object position distributions.
    % \item (Not sure if we should include this:) Is a sample efficient and lightweight transfer framework...does not require training on massive internet scale datasets or heavily randomized prior domains, but instead relies on a modest amount of data in the target domain (on a different task).
\end{enumerate}

In the few-shot setting, on long-horizon multi-step real-world tasks, these advantages enable \ouralgo~to outperform prior SOTA methods in sim2real and vision representation learning by $25$-$40\%$. To our knowledge, this is the first work that shows that using language to learn a domain-invariant visual representation can help improve the sample efficiency and performance of sim2real transfer.

\section{Related Work}
\label{sec:related-work}
Our main contribution is a method to learn domain-invariant image representations by exploiting natural language descriptions as a bridge between domains for sim2real transfer. While we believe this has not been explored before, significant related research has been done in vision-language pretraining, sim2real techniques, and domain-invariant representations for control. 

% Prior work in domain-invariant representation learning for domain adaptation in robotics has also proven effective, but they are primarily confined to the visual modality without using language as a bridge between representations of different domains.

\subsection{Vision Pretraining for Robotics} Various works have found that \textbf{vision-only pretraining} improves performance on image-based robotic policies. 
% In general, these follow the pretrain-finetune approach just like our method, where an image encoder is first pretrained, and then during BC or RL, the image backbone is usually frozen, and an added policy head is allowed to train.
Prior work has explored pretraining objectives ranging from masked image modeling~\citep{radosavovic2023real}, image reconstruction~\citep{Zhao2022WhatMR, gupta2022maskvit, seo2023masked}, contrastive learning~\citep{laskin2020curl, he2020momentum}, video frame temporal ordering~\citep{jing2023exploring}, future frame prediction~\citep{Zhao2022WhatMR}, and image classification~\citep{yuan2022pre, wang2022vrl3} on internet-scale datasets such as ImageNet~\citep{deng2009imagenet}, Ego4D~\citep{grauman2021ego4d}, Something-Something~\citep{goyal2017something}, and Epic Kitchens \citep{Damen2018EPICKITCHENS}. While these vision-only pretraining objectives learn good representations for robotic control within a specific domain distribution (such as the real world), they are not necessarily robust to the wide domain shifts encountered during sim2real.

In \textbf{vision-language pretraining}, contrastive learning~\citep{radford2021learning, zhai2021lit} has been shown to learn valuable representations for robotic tasks~\citep{shridhar2021cliport, shridhar2022peract}. However, these pretrained visual representations are often overly influenced by the semantics of language captions. This results in a representation that is too object-centric to differentiate between different frames of a robot demonstration, lacking the level of granularity needed for spatial-temporal understanding. R3M~\citep{nair2022r3m} addresses this by learning semantics from language labels of videos but also training with a time contrastive loss between video frames.
% While our pretraining approach is focused on language and does not have a temporal aspect to it, our approach can be combined with an R3M-styled loss.
Prior work in multimodal representations~\citep{zhu2023languagebind} found language to be effective in aligning representations learned across multiple modalities including depth and audio. Instead of using language to bridge modalities, our approach uses language to bridge visual representations between domains.
% By using language specifically describing spatial relations between objects and the robot in the scene, our approach learns domain-invariant features that focus on the important aspects of an image observation needed for low-level control.

\subsection{Sim2Real}
While we approach sim2real through vision-language pretraining, there are many alternative, well-researched techniques. \textbf{Domain randomization}~\citep{andrychowicz2020learning, matas2018sim, tobin2017domain} involves varying physical parameters and visual appearances of the simulation to train a policy that functions in a wide distribution of domains that hopefully also covers the target domain distribution. However, domain randomization requires a large amount of diverse training data and attempts to be simultaneously performant in an overly broad distribution of states, leading to a suboptimal and conservative policy that takes longer to train. \textbf{System identification}~\citep{yu2017preparing, kaspar2020sim2real} involves tuning the simulation parameters to match the real world in order to create a custom-tailored simulation environment that easily transfers to the real domain. However, this process is very engineering intensive and time consuming, and it may be intractable to simulate all real world physical interactions with high fidelity and throughput. In contrast, our sim2real approach can handle large source and target domain discrepancies with a few target task demonstrations and does not require system identification or domain randomization.

\subsection{Domain-Invariant Representations}
Several methods have been proposed to learn domain invariant representations. The domain-adaptation community has extensively researched using \textbf{Generative Adversarial Networks (GANs)} to map images from one distribution into another, using pixel space as a medium of common representation~\citep{james2019sim, bousmalis2017unsupervised, ho2021retinagan, rao2020rl}. However, GANs require a large training dataset and are notorious for unstable training. Additionally, enforcing similarity on the input image side at the pixel level is less efficient than our method, which encourages cross-domain distributional similarity in a compact, low-dimensional image encoder space. Furthermore, researchers in self-driving have studied using \textbf{semantic segmentation} and depth maps~\citep{muller2018driving, ai2023invariance} as a common representation space between domains, though their effectiveness has only been demonstrated in navigation tasks with binary segmentation masks, which is too simplified for the long-horizon manipulation tasks we consider.

\subsection{Language and Robotics}
A growing body of work has investigated training \textbf{multitask robotic policies} conditioned on language instruction embeddings ~\citep{jang2021bc, lynch2021language, calvin2021, hulc22, shao2020concept, sodhani2021multi, silva2021langcon, karamcheti2021lila}, or a combination of language instructions and goal images/demonstrations ~\citep{jiang2022vima, shah2023mutex, yu2022deltaco}. Our approach also involves learning a language-conditioned policy, but unlike prior work, our main novelty is using language for a second use-case: as scene descriptors during pretraining to pull together semantically similar image observations between two visually dissimilar domains. Language has also been used for \textbf{reward shaping} in RL ~\citep{nair2021lorel, goyal2019learn, goyal2020pixl2r, fan2022minedojo, ma2023liv}, and as a high-level planner in long-horizon tasks ~\citep{huang2022inner, saycan2022arxiv, chen2022nlmapsaycan, raman2023cape}. These areas of research are more ancillary to our contributions, as we demonstrate our approach with IL instead of RL and with fine-grained manipulation tasks that do not require extensive planning.

\section{Problem Description}
\label{sec:problem-setup}
In this work, we address the problem of few-shot visual imitation-learning (IL): learning a visuomotor manipulation policy in the real world based on a few real-world demonstrations. We assume access to a large amount of simulation data and \rev{cast} few-shot IL as a sim2real problem. More concretely, we render the few-shot IL problem as a $k+1$ multi-task IL problem: $k$ tasks from simulation and the target task (with a few demonstrations) in the real world. In general terms, we assume a \textit{source domain} in which data can be acquired cheaply and a \textit{target domain} where data is expensive to collect.

In our setting, we consider access to two datasets across two domains: $\priorD$, which spans multiple tasks in the source domain, and $\targetDtargetT$, which contains a small number of demonstrations of the target task in the target domain we want to transfer to. Thus, we assume that $|\priorD| >> |\targetDtargetT|$, due to how expensive target domain data collection is (such as in the real world). We make two assumptions about the two domains. First, we assume the source and target tasks are all of the same general structure, such as multi-step pick-and-place task compositions, but with different objects and containers across different subtasks. Otherwise, transfer would be infeasible in the low-data regime if the source and target domain tasks lack similarity. Second, to train a common policy for both domains, we assume the domains share state and action space dimensionality. We make no further assumptions about the similarity between the two domains.

All of our datasets are in the form of expert trajectories. Each trajectory, $\tau = \{(I_t, s_t, [a_t, l_t], l_{task} )\}$, is a sequence of tuples containing an image observation, $I_t$ ($128\times128$ RGB), robot proprioceptive state, $s_t$ (end effector position and joint angles), and a language instruction of the task, $l_{task}$. Note that $l_{task}$ is the same over all timesteps of all trajectories in a given task. $[a_t, l_t]$ denotes that a trajectory may optionally also include robot actions (in which case we consider the trajectory a full demonstration) and/or a language description of the image $I_t$. In the following sections, we identify with $\tau[L]$ a trajectory with language descriptions $l_t$, but no actions $a_t$. Similarly, $\tau[A]$ is a full demonstration with actions, $a_t$, but no language descriptions, $l_t$.

The language labels for images \rev{can be automatically generated from} a programmatic function that maps image observations to language scene descriptions depending on the relative position between the robot and the objects in the scene. \rev{We elaborate on these language labels and how to automatically collect them in Section~\ref{sec:method:langlabeling}.} Note that these language \textit{scene descriptions}, $l_t$, are different from the language \textit{instruction} associated with each task, $l_{task}$.

% We assume that $\priorD$ contains demonstration trajectories with both action labels and language descriptions in each transition $\tau[AL]$, $\targetDpriorT$ contains trajectories with image language labels $\tau[L]$, and $\targetDtargetT$ contains demonstrations with action labels and no language descriptions $\tau[A]$.
Different data elements and types of trajectories will be used during pretraining and policy few-shot training: during pretraining, we use $\tau[L]$ image-language $(I_t, l_t)$ pairs from $\priorD \cup \targetDtargetT$. During policy learning, we use $\tau[A]$ data: $(\rev{I_t}, s_t, a_t, l_{task})$ tuples from $\priorD \cup \targetDtargetT$. In the next section, we explain how these two steps are defined for \ouralgo{}.

\section{\ouralgo: Few-Shot IL with Sim\&Real}
\label{sec:method}
\begin{figure}[t] 
    \centering
    \includegraphics[width=1\columnwidth]{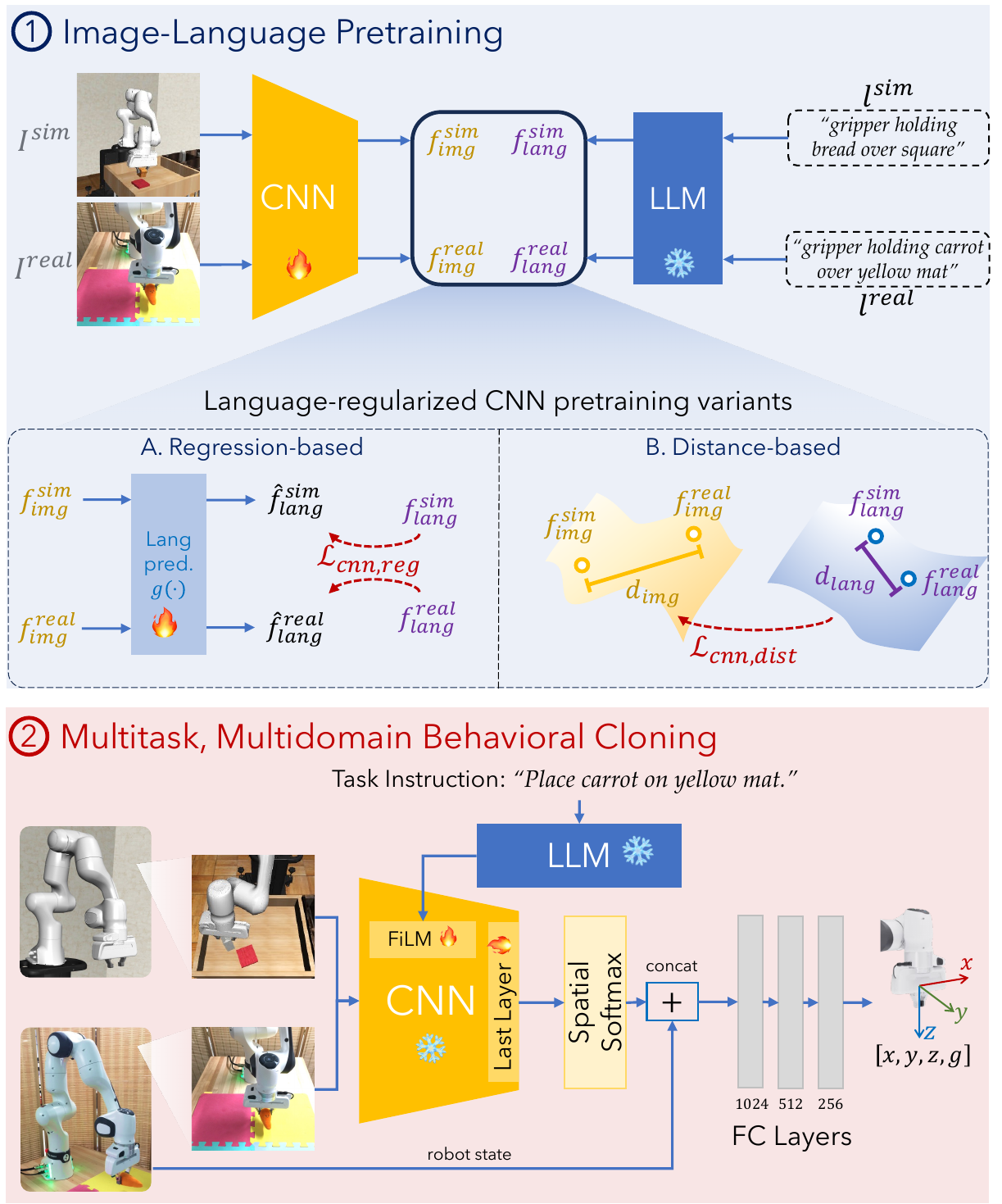}
    \caption{\textbf{Method.} (i) Top: During \textbf{Image-Language Pretraining}, we train the image encoder $f_{cnn}$ using the language embeddings associated with descriptions of both sim and real image observations. $f^{d}_{img}$ and $f^{d}_{lang}$ refer to the output features of the CNN and the LLM, respectively, in domain $d$. With regression-based loss (A) the image embeddings are pushed to predict the corresponding language embeddings  whereas with distance based loss (B) the pair of image embeddings is pushed together/apart based on the similarity of the language embeddings.
    (ii) Bottom: During \textbf{Multitask, Multidomain BC}, we freeze our pretrained $f_{cnn}$, add adapter modules and a policy head and allow the last layer of the CNN to finetune, then train the resulting multitask language-conditioned policy on $\priorD \cup \targetDtargetT$.}
    \label{fig:method} 
\end{figure}

In our method, we adopt the common \textit{pretrain-then-finetune} learning paradigm (see Fig.~\ref{fig:method}). First, we pretrain an image backbone encoder on cross-domain language-annotated image data (Sec.~\ref{sec:method:imglangpt}). Then, we freeze this encoder and train a policy network composed of trainable adapter modules and a policy head to perform behavioral cloning (BC)~\cite{schaal1999imitation} on action-labeled data from both domains (Sec.~\ref{sec:method:bc}). To leverage the simulation data, we train a $k+1$ multi-task BC policy that learns for $k$ tasks in the source domain (sim) and $1$ in the target domain (real, few shot).

\subsection{\rev{Automatic Language labeling of Images}}
\label{sec:method:langlabeling}
\rev{To acquire image-language pairs for pretraining, we implement an automated pipeline for labeling the images of a trajectory that occurs synchronously during scripted policy demonstration collection (see Appendix~\ref{appendix:scripted-policies}). Each if-case in the scripted policy corresponds to a stage index, where in pick-and-place, the first stage corresponds to the gripper moving to a point above the object, the second stage corresponds to the gripper moving vertically down toward the object, and so on. We define a list of template strings describing the scene for each of these stages, so the stage indexes into the template string list, giving us our language annotation for the image. See Table~\ref{tab:lang-templates} in the Appendix for all template strings, and Appendix~\ref{appendix:lang-labeling} for details about our language labeling procedure.}

\rev{However, our language labeling process need not be synchronously coupled with scripted policy demonstration collection. We also implemented a labeling process using off-the-shelf vision-language models to detect the location of objects and the gripper in the image to predict the stage number. This process can be run on previously-collected trajectories and requires only the images of a trajectory alone, without need for additional action or state information. We describe this second process in Appendix~\ref{appendix:lang-labeling-vlm}. Empirically, using language from this second, more scalable automated approach does not degrade the performance of our method.}

\subsection{Cross-Domain Image-Language Pretraining}
\label{sec:method:imglangpt}

\rev{After collecting trajectories with language labels,} our first step in \ouralgo{} involves learning a domain-invariant representation that will enable leveraging simulation data for few shot IL. For that, we need to learn an image observation encoder, $f_{cnn}: \rev{I_t} \rightarrow \mathbb{R}^{d_{cnn}}$, that attains the following property:
% \begin{enumerate}[wide, labelwidth=!, labelindent=0pt]
%     \item \textbf{Cross-domain scene similarity}: 
    it should preserve the semantic similarity of scenes in images between the two domains. For instance, if both \rev{image $I^{s}$} from $\priorD$ (sim) and \rev{image $I^{t}$} from $\targetDtargetT$ (real world) show the robot's gripper open and a few inches above the object to grasp, even if from different viewing angles, then we want their image embeddings to be close together in the learned image encoding space. This will facilitate policy learning later, as the policy will need to draw from a similar distribution of actions for similar scene semantics, which are now already mapped into similar visual features.
    % \item \textbf{Fine-grained representations for low-level control}: Our learned representation should avoid over-optimizing for scene similarity, else resulting image embeddings would only reflect the (coarse) high-level semantics ---e.g., is the gripper open? is it above the object?--- and ignore low-level details such as how far and in what direction is the gripper from the object, which are critical to low-level control.
% \end{enumerate}

Theoretically, off-the-shelf pretrained vision-language models (VLMs)~\citep{radford2021learning, nair2022r3m} should already possess these properties as they were trained on a massive distribution of image and language data. However, in the context of robot manipulation, pretrained VLMs tend to encode all observations of the trajectory into a very narrow region of the embedding space without sufficient distinction for task-relevant, semantic aspects of the image such as the location of the gripper in relation to the manipulated objects. This renders them unsuitable without additional finetuning for our application (see Sec.~\ref{sec:experimental-results}).

In \ouralgo{}, we propose an alternative approach to obtain a visual representation with the aforementioned desired property. We train a ResNet-18~\citep{he2015resnet} from scratch as our image encoder using image-language tuples $(I^{s}, l^{s})$ from $\priorD$ and $(I^{t}, l^{t})$ from $\targetDtargetT$. We denote this vision language pretraining dataset as $\mathcal{D}_{VL} = \{ (I^d, l^d): (I^d, l^d) \in \priorD \cup \targetDtargetT \}$, where $d$ is either the source or target domain. The images are observations collected during 100 demonstrations from each of the tasks in $\priorD$ and 25-100 demonstrations from $\targetDtargetT$, totaling around 10k images per domain. We assume that the two sets of language descriptions in $\priorD$ and $\targetDpriorT$ are similarly distributed; otherwise, language may not help learn domain-invariant features between $\priorD$ and $\targetD$.
% By training on a narrow distribution of data, we enable our representation to learn subtle differences between images of a trajectory (property 2) while also picking out on the semantics of scenes in cross-domain images via language (property 1).

To effectively leverage language as a bridge between visually different domains, we need a well-tuned (frozen) language model, $f_{lang}: l \rightarrow \mathbb{R}^{d_{lang}}$, to map strings to $d_{lang}-$dimensional language embeddings. We use off-the-shelf miniLM~\citep{wang2020minilm}, since prior work~\citep{hulc22} has demonstrated its effectiveness for language-conditioned control policies compared to other small, off-the-shelf language models.

Given the data and the language embedding described above, we propose two variants in \ouralgo{} for the image-language pretraining step that can obtain a sim-real agnostic representation based on language supervision (see Fig.~\ref{fig:method}(i)A-B):

\subsubsection{\textbf{Language-Regression}}\label{sec:method:lang-reg}
Our first variant is a straightforward use of language supervision to shape the image embedding space: predicting the language embedding of the description, $l^d$, given the embedding of the corresponding image, $I^d$. We sample image-language pairs from the $\mathcal{D}_{VL}$ \rev{dataset} defined above: $(I^d, l^d) \sim \priorD \cup \targetDtargetT$. Let $g: \mathbb{R}^{d_{cnn}} \rightarrow \mathbb{R}^{d_{lang}}$ be a single linear layer (language predictor in Fig.~\ref{fig:method}(i)(A)) trained to minimize the following loss:
\begin{equation}
    \mathcal{L}_{cnn, reg}(\mathcal{D}_{VL}) = \big\| g \left( f_{cnn}(I^d) \right) - f_{lang}(l^d) \big\|_2^{2}
\end{equation}
We use the loss to train both the language predictor and the CNN backbone. The loss provides strong language supervision by encouraging $f_{cnn}$ to directly regress toward the frozen language embeddings of the image descriptions, effectively making the pretrained image encoder reflect the LLM embedding space.

\subsubsection{\textbf{Language-Distance Learning}}
\label{sec:method:lang-dist}
We also experiment with a second variant of image-language pretraining that incorporates language with a softer form of supervision. We posit that the exact values of the language embeddings do not themselves convey meaning; rather, key information about the semantic similarity of two images lies in the pairwise distances between their corresponding two language embeddings. Thus, we design an objective to regress the image embedding distances between a pair of images from the two domains to their corresponding language distance:
\begin{equation}
    \mathcal{L}_{cnn, dist}(\mathcal{D}_{VL}) = \big\| f_{cnn}^{\top}(I^s) f_{cnn}(I^t) - d \big(l^s, l^t \big) \big\|_2^{2}
\end{equation}
where the language distance function we use, $d: l \times l \rightarrow \mathbb{R}$ is BLEURT~\citep{sellam2020bleurt}, a learned similarity score between two strings commonly used in the NLP community. We normalize $d(\cdot, \cdot)$ between 0 and 1 for all possible $\left( l^{s}, l^{t} \right)$ pairs in our image-language dataset, where $1$ indicates the highest similarity between any two strings in the dataset. Empirically, over our set of language descriptions, we found BLEURT provided a richer signal than simply taking dot products or $\ell$-2 distances between language embeddings. The output of $f_{cnn}$ is unit normalized before taking the dot product. We compare both variants (see Sec.~\ref{sec:experimental-results}) to assess whether the additional degrees of freedom from the looser distance supervision are beneficial later on for policy training.

\subsection{Multitask, Multidomain Behavioral Cloning}
\label{sec:method:bc}
Our second step in \ouralgo{} involves learning a multi-domain, multi-task, language-conditioned BC policy (see Fig.~\ref{fig:method}(ii)). By leveraging our learned domain-invariant representation for robotic control, this policy should be able to perform well in real-world task with only a few demonstrations, thanks to the additional information it can extract from simulation.

% we use our pretrained $f_{cnn}$ as the basis of our a multi-domain, multi-task, language-conditioned BC policy (see Figure~\ref{fig:method}(b)). 
During this phase of policy learning, we freeze all but the last layer to preserve the semantic scene information encoded in the learned, domain-invariant representation, $f_{cnn}$, while enabling the network to adapt to the new downstream task of low-level control. We also insert trainable FiLM layer blocks~\citep{perez2018film} as adapter modules in $f_{cnn}$ to process the language instruction embeddings between the frozen convolution layers. Finally, we include a few fully-connected layers as a policy head to process the image feature, $f_{cnn}(I_t)$, and proprioceptive state, $s_t$, and train the resulting policy $\pi$ with BC loss to predict the mean and standard deviation of a multivariate Gaussian action distribution, as described below.

Let our training dataset $\mathcal{D}_{BC} = \priorD \cup \targetDtargetT$ be a set of demonstrations $\tau^d$, for domain $d \in \{ \text{source, target} \}$. As explained in Sec.~\ref{sec:problem-setup}, each demonstration is a sequence of tuples $x_t = \left( I_{t}^{d}, s_{t}^{d}, a_{t}^{d}, l_{task} \right)$ containing the image observation, proprioceptive state, language instruction for the task, and action at timestep $t$. We train with the following standard BC negative log probability loss~\citep{pomerleau1988alvinn}:
\begin{equation}
\label{eqn:policyloss}
    \mathcal{L}_{\pi} ( \mathcal{D}_{BC} ) = \frac{1}{B}\sum\limits_{\substack{x_t \sim \tau^{d} \\ \tau^{d} \sim \mathcal{D}_{BC}}}-\log{\pi\left(a_{t}^{d} \big| f_{cnn}(I_{t}^{d}), s_{t}^{d}, l_{task} \right)}
\end{equation}
where $B$ denotes the batch size.

% After training $f_{cnn}$, we add trainable adapter modules and a policy head to create a policy network $\pi: f_{cnn}(o_t) \times s_t \times z_{task} \rightarrow a_t$ that predictions actions given image observation embeddings, proprioceptive state, and language instruction embedding $z_{task} = f_{lang}(l_{task})$. Additionally, we allow the final layer of $f_{cnn}$ to continue training during this stage.

% Note that this is the second form of language we use in our approach; $z_l$ are embeddings of language instructions that describe the task, not the language descriptions $l$ describing scenes that was used in the previous pretraining phase.
% We collect several thousand scripted-policy trajectories across $T^{(1)}_{1:k}$ (where the observations $o_t$ were also used in Phase 1 without the action labels) and $<100$ target task $T^{(2)}$ trajectories. We also assume that the language instructions between tasks in $T^{(1)}_{1:k}$ and $T^{(2)}$ are of the same class of tasks (such as multistep pick-and-place).

% We train over $k+1$ different tasks in both domains: $T^{(1)}_{1:k} \cup T^{(2)}$. In each batch, we sample uniformly at random from $m$ of the $k+1$ tasks, and then query the dataset for a fixed number of transitions in the trajectory for each of the $m$ selected tasks.
The policy is trained on $k+1$ tasks: $k$ from $\priorD$ (thousands of trajectories per task) and $1$ from $\targetDtargetT$ ($\leq 100$ trajectories, see Sec.~\ref{sec:experimental-setup}). In each batch, we sample $m$ tasks uniformly at random from the $k+1$ tasks, and then query $\mathcal{D}_{BC}$ for a fixed number of transitions from trajectories for each of the $m$ selected tasks.

We hypothesize that cross-domain image-language pretraining (Sec.~\ref{sec:method:imglangpt}) improves policy learning because it helps ensure that image observations of different domains depicting semantically similar scenes map into similar regions of the learned embedding space. 
% This representation encourages the learned policy to predict semantically similar action distributions given similar images across different domains.
This accelerates learning not only on $\priorD$ data but also helps alleviate data scarcity in $\targetDtargetT$ because the pretrained image backbone encodes $\targetDtargetT$ images into an in-distribution region of the learned image embedding space, alleviating common issues with visual distribution shift and enabling our method to leverage simulation data to compensate for the lack of real-world action-labeled data, improving sim2real transfer.

\section{Experimental Setup}
\label{sec:experimental-setup}
We evaluate \ouralgo{} in two settings: a \texttt{sim2sim} setting where we test the transfer abilities between two simulated domains with visual and physical differences, and the \texttt{sim2real} setting, where the few shot IL is defined in the real world and we use simulation to address the data scarcity. \texttt{Sim2sim} serves as a platform to evaluate in depth the behavior of \ouralgo{} with a fully controlled domain gap, while \texttt{sim2real} is our setting of interest for this work. We will use three task suites that we explain below. See Figure~\ref{fig:film-strips} in the Appendix for detailed frame rollouts of each task. In a slight overload of notation from Sec.~\ref{sec:problem-setup}, here we use $\priorD$ and $\targetD$ to denote the source and target domains, respectively. 

\subsection{\texttt{Sim2Sim} and \texttt{Sim2Real} Environment Differences}
% \label{sec:exp:sim2simdiff}
\label{sec:exp:sim2sim-sim2real-diff}
In \texttt{sim2sim}, $\priorD$ and $\targetD$ are both sim environments with large differences in camera point-of-view (third person vs. first person), joint friction, and damping.
% \begin{enumerate}
%     \item \textbf{Camera point-of-view}: $\priorD$ image observations are third person (looking toward the robot), and $\targetD$ image observations are first person (over the shoulder), a large viewing angle change.
%     \item \textbf{Friction and Damping}: Joint friction and damping coefficients are $5 \times$ and $50\times$ higher in $\targetD$ than $\priorD$, which changes significantly the dynamics.
% \end{enumerate}
% \subsection{\texttt{sim2real} Environment Differences}
% \label{ss:s2rdiffs}
In the \texttt{sim2real} setting,
% in contrast with prior work that requires significant engineering to reduce differences between the sim $\priorD$ and real $\targetD$ environment,
we employ a setup with a wide sim2real gap that we aim to bridge using language that includes differences in control frequency, task goals, visual observation appearance, objects, and initial positions.
% \begin{enumerate}
%     \item \textbf{Control frequency}: The simulated $\priorD$ policy runs at 50Hz while the real world $\targetD$ policy runs at 2Hz.
%     \item \textbf{Objects}: The objects on the scene in each task differ between simulation and real data, except the robot itself.
%     \item \textbf{Visual Observation}: Backgrounds and camera angles are markedly different between the two domains.
%     \item \textbf{Initial positions}: The initial object and robot positions are different across sim and real.
%     % \item \textbf{Action magnitude}: Real robot x, y, and z action are restricted so each are $\le 1$, and higher values are scaled before being executed. 
% \end{enumerate}
More details between the two environments in \texttt{sim2sim} and \texttt{sim2real} can be found in Appendix~\ref{appendix:sim2sim-sim2real-diff}.

\subsection{Evaluation Metrics}
\label{sec:exp:metric}
For all \texttt{sim2sim} and \texttt{sim2real} experiments, we measure task success rate. In \texttt{sim2real}, this is calculated through ten evaluation trials for each of two seeds per task, for a total of 20 trials per table entry. In each set of ten trials, we place the object in the same ten initial positions and orientations, evenly distributed through the range of valid initial object positions. In \texttt{sim2sim}, we also run two seeds per setting and take a success rate averaged over 720 trials between the two seeds in the final few hundred epochs of training.
% and average the success rate on twelve evaluation trial every ten epochs between 300 and 600 epochs (for a total of 720 trials).

\subsection{Data}
\subsubsection{Environments} For each of our tasks, we design simulation environments on top of Robosuite~\citep{zhu2020robosuite} in Mujoco~\citep{todorov2012mujoco}. For the real environment, we use Operational Space Control~\cite{khatib1987unified} to control the position of the end-effector of the robot in Cartesian space. In both simulation and real, we work with a 7-DOF Franka Emika Panda arm and use a common action space consisting of the continuous xyz delta displacement and a continuous gripper closure dimension (normalized from $[-1, 0]$).  The robot proprioception space is 22-dimensional, consisting of the robot's xyz end-effector position, gripper state, and sine and cosine transformations of the 7 joint angles. The image observation space is $128 \times 128$ RGB images.

\subsubsection{Overview of Tasks} For each task suite, we collect data from simulated domain $\priorD$ and real target domain $\targetD$ (for \texttt{sim2real}) or sim target domain $\targetD$ (for \texttt{sim2sim}). All demonstrations in sim and real are collected with a scripted policy (see Appendix for further details). Sim trajectories range from 200-320 timesteps long, operated at \SI{50}{\hertz}, while real trajectories run at \SI{2}{\hertz} and range from 18-45 timesteps. Our three task suites allow us to test the effectiveness of \ouralgo{} for sim2real in a wide variety of control problems ranging from simple stacking in task suite 1, to multi-step long-horizon pick and place in task suite 2, to deformable, hard-to-simulate objects in task suite 3.

\begin{figure}[t!]
\centering
\includegraphics[width=0.32\linewidth]{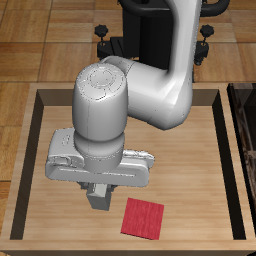}%
\hfill%
\includegraphics[width=0.32\linewidth]{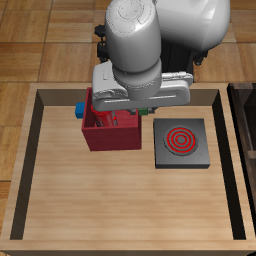}%
\hfill%
\includegraphics[width=0.32\linewidth]{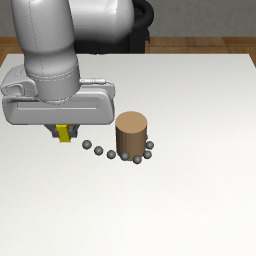}\\
\vspace{5pt}
\includegraphics[width=0.32\linewidth]{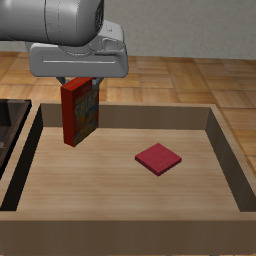}%
\hfill%
\includegraphics[width=0.32\linewidth]{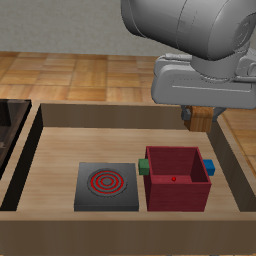}%
\hfill%
\includegraphics[width=0.32\linewidth]{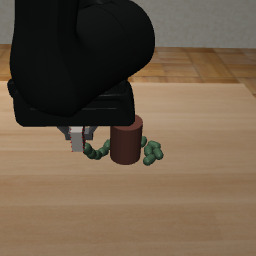}\\
\vspace{5pt}%
\includegraphics[width=0.32\linewidth]{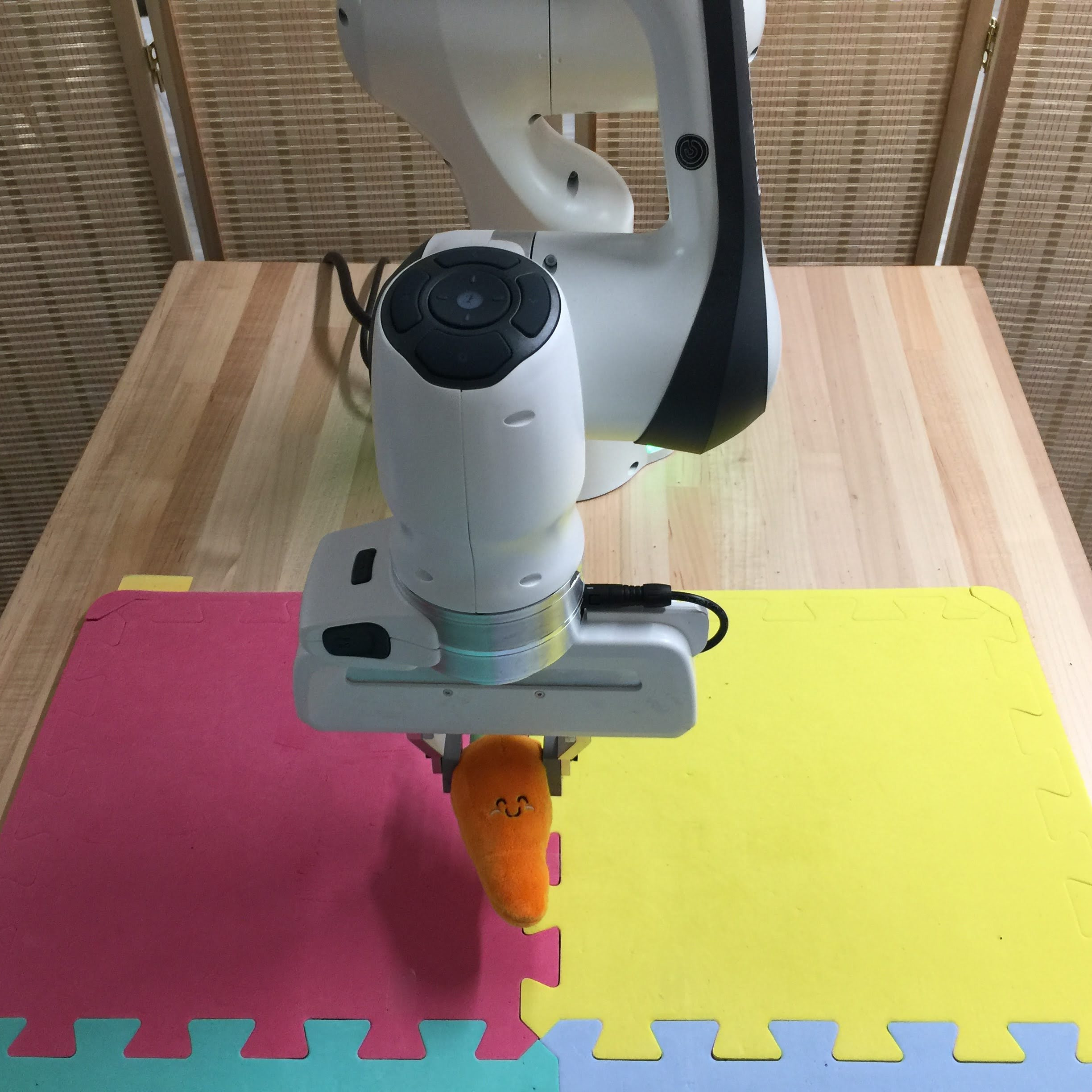}%
\hfill%
\includegraphics[width=0.32\linewidth]{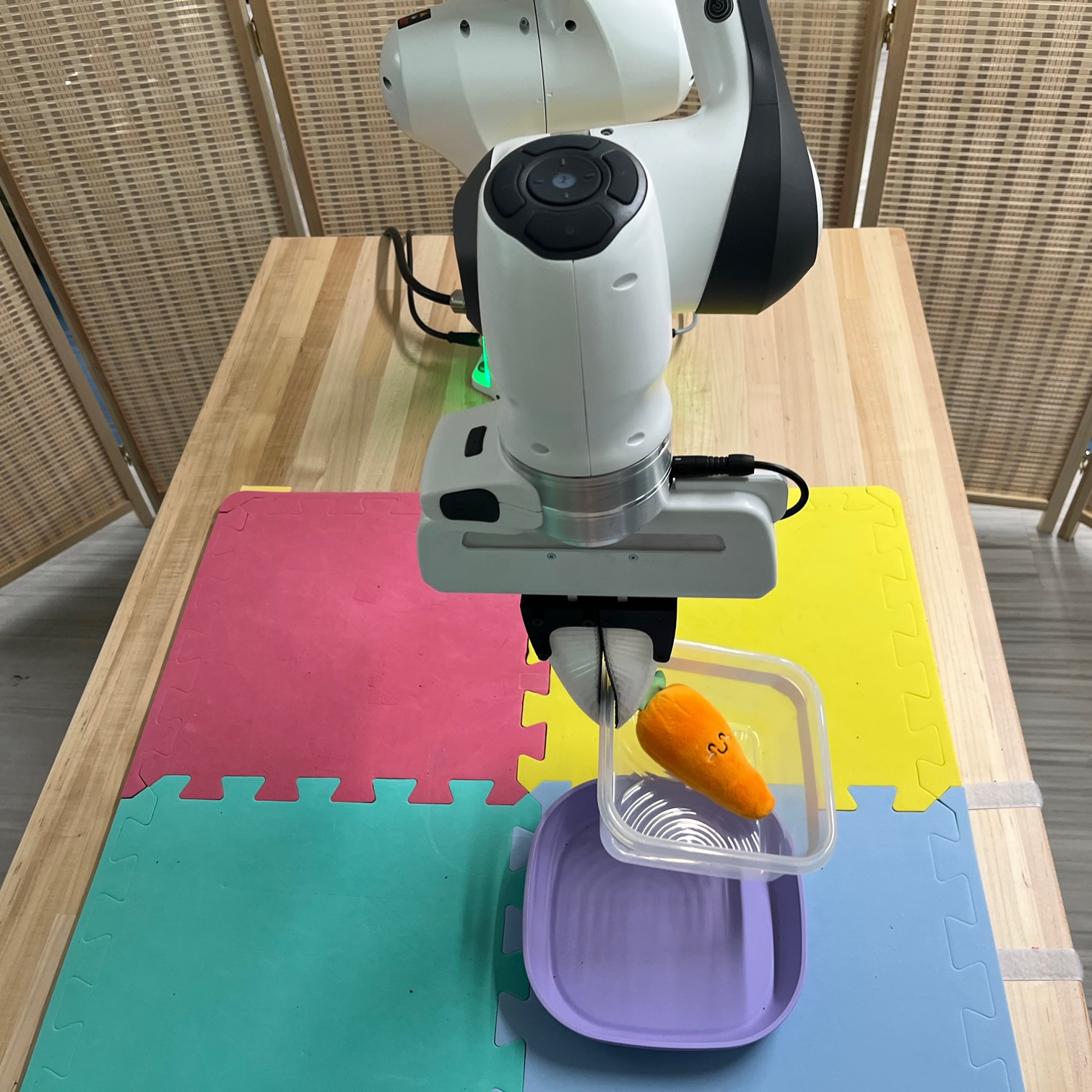}%
\hfill%
\includegraphics[width=0.32\linewidth]{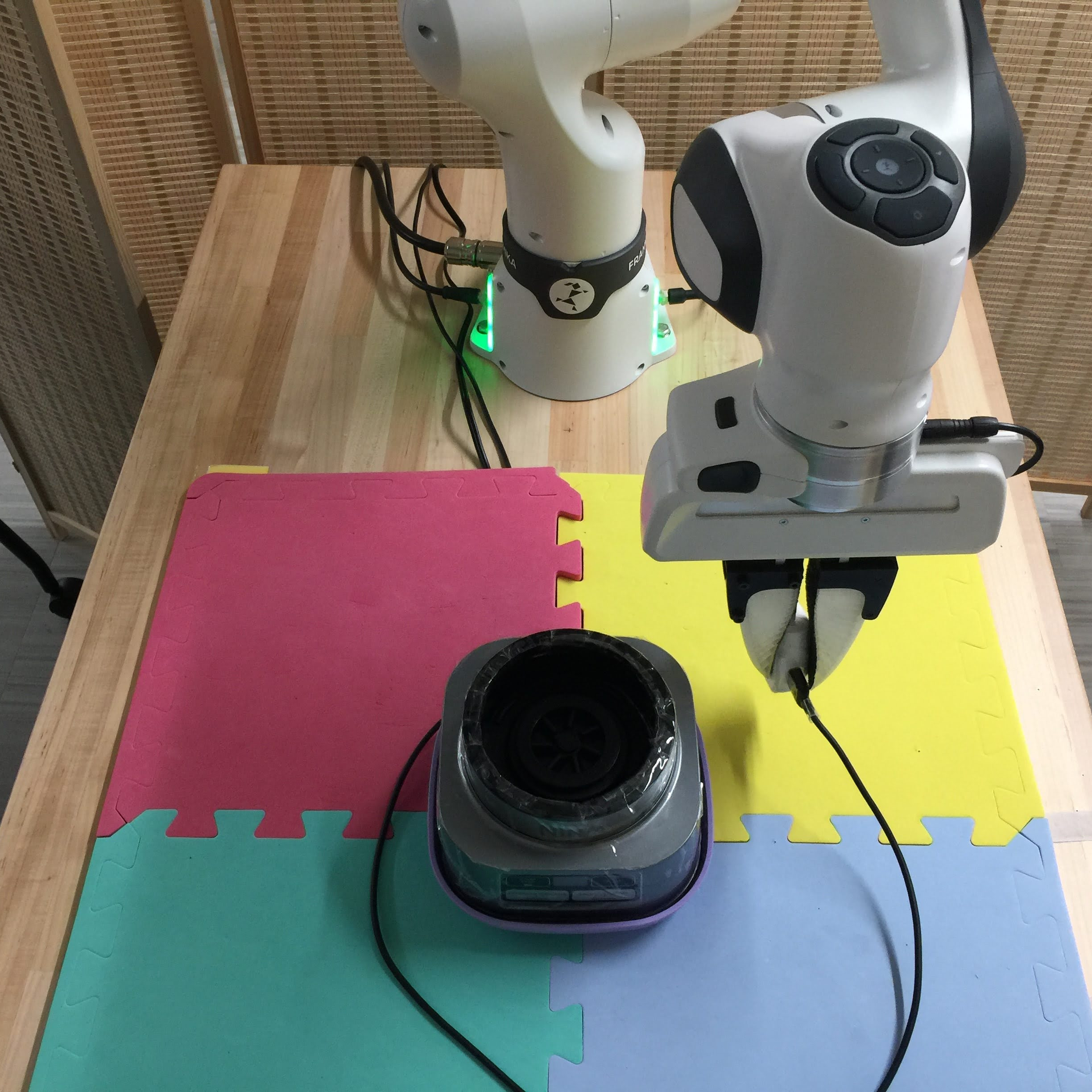}\\
\caption{The columns depict the three task suites while each row represents an image domain. \textit{\rev{Rows from Top to Bottom}:} Simulation $\priorD$, \texttt{sim2sim} $\targetDtargetT$, \texttt{sim2real} $\targetDtargetT$\rev{.} \textit{\rev{Columns from} Left to Right:} Stack Object, Multi-step Pick and Place, and Wrap Wire tasks. While similar enough to transfer prior knowledge between them, our $\priorD$ and $\targetD$ task versions \rev{have} a considerable gap (Sec.~\ref{sec:exp:sim2sim-sim2real-diff}) that we are able to bridge using language as regularization for the image representations.}  
\label{fig:tasks}
\end{figure}

%%%%%%%%%%%%%%
%%% TASK 1 %%%
%%%%%%%%%%%%%%
\subsection{Task Suite 1: Stack Object}
\label{sec:exp:stack}
In our first suite of tasks, the robot must move an object to a target. In the simulated domain $\priorD$, the target is on top of a wooden coaster, and there are four objects: milk carton, soda can, bread, and cereal box, which correspond to the four tasks. Both the object and coaster positions are randomized over the entire workspace. We collect and train on $400$ demonstrations per task ($1600$ total) as our $\priorD$ simulation data.
\begin{table*}[htbp]
  \centering
  \caption{\texttt{sim2real}: Performance by number of real world trajectories}
  \resizebox{\textwidth}{!}{
  \begin{tabular}{ccccccccccccccc}
    \toprule
    Method & \multicolumn{2}{c}{Action-labeled Data} &  \multicolumn{3}{c}{Stack Object} & \multicolumn{6}{c}{Multi-step Pick and Place} & \multicolumn{3}{c}{Wrap Wire}\\
    \cmidrule(r{4pt}){2-3} \cmidrule(l){4-6} \cmidrule(l){7-12} \cmidrule(l){13-15}
    & Sim & Real & \multicolumn{3}{c}{Success Rate \rev{(\%)}} & \multicolumn{3}{c}{Success Rate \rev{(\%)}} & \multicolumn{3}{c}{Subtasks Completed} & \multicolumn{3}{c}{Success Rate \rev{(\%)}} \\
    \cmidrule(l){4-6}\cmidrule(l){7-9}\cmidrule(l){10-12} \cmidrule(l){13-15}
    &  $\priorD$ &  $\targetDtargetT$ &  25 & 50 & 100 & 25 & 50 & 100 & 25 & 50 & 100 & 25 & 50 & 100 \\
    \midrule 
    
    \rev{No Pretrain ($\targetD$)} & -- & \ding{51} & 20 & 30 & 45 & 0 & 30 & 35 & 0.45 & 1.05 & 1.05 &  20 & 15 & 45  \\
     \rev{No Pretrain ($\priorD + \targetD$)} & \ding{51} & \ding{51} & 35 & 20 & 55 & 45 & 25 & 55 & 1.15 & 1.0 & 1.4 & 25 & 20 & 20 \\
    \midrule
    \rev{MMD} & \ding{51} & \ding{51} & 25 & 35 & \textbf{80} & 20 & 10 & 35 & 0.8 & 0.9 & 1.1 & 5 & 10 & 20 \\
    \rev{Domain Random.} & \ding{51} & \ding{51} & 40 & 60 & 40 & 10 & 10 & 25 & 0.7 & 0.6 & 0.7 & 0 & 0 & 0 \\
    \rev{ADR+RNA} & \ding{51} & \ding{51} & 35 & 30 & 35 & 15 & 25 & 40 & 0.85 & 0.8 & 1.3 & 0 & 10 & 0 \\
    \midrule
    \rev{Lang Reg. (ours)} & \ding{51} & \ding{51} & 40 & \textbf{75} & \textbf{80} & \textbf{60} & \textbf{80} & \textbf{90} & \textbf{1.45} & \textbf{1.8} & \textbf{1.9} & \textbf{45} & \textbf{40} & 45\\
    \rev{Lang Dist. (ours)} & \ding{51} & \ding{51} & \textbf{60} & 45 & \textbf{80} & 55 & 70 & 75 & 1.35 & 1.65 & 1.6 & 30 & 25 & \textbf{75}  \\
    \rev{Stage Classif.} & \ding{51} & \ding{51} & 40 & 60 & 60 & 50 & 60 & 50 & \textbf{1.45} & 1.55 & 1.5 & 30 & \textbf{40} & 50 \\
    \midrule
    \rev{CLIP (frozen)} & \ding{51} & \ding{51} & 25 & 5 & 15 & 10 & 15 & 40 & 0.3 & 0.45 & 1.0 & 35 & 35 & 30 \\ 
    \rev{R3M (frozen)} & \ding{51} & \ding{51} & 30 & 45 & 65 & 15 & 60 & 55 & 0.7 & 1.4 & 1.5 & 5 & 25 & 25 \\

    \bottomrule
  \end{tabular}
  }
  \label{tab:real}
\end{table*}

\begin{table*}[htbp]
  \centering
  \setlength{\tabcolsep}{0.2em}
  \caption{\texttt{sim2sim}: Success Rate by Task (\%)}
  \resizebox{\textwidth}{!}{
  \begin{tabular}{ccccccccccccccc}
    \toprule
     
     Pretraining & \multicolumn{5}{c}{Stack Object} & \multicolumn{5}{c}{Multi-step Pick and Place} & Wrap Wire \\
    \cmidrule(l){2-6}
    \cmidrule(l){7-11}
    \cmidrule(l){12-12}
    & 1 & 2 & 3 & 4 & avg & 1 & 2 & 3 & 4 & avg & 1\\
    \midrule
    None ($\targetD$ data only) & $15.2 \pm 6.5$ & $18.9 \pm 6.7$ & $31.9 \pm 8.5$ & $25.4 \pm 9.2$ & $22.9$ & $20.8 \pm 7.5$ & $17.7 \pm 4.4$ & $16.3 \pm 5.0$ & $17.3 \pm 8.1$ & $18.0$ & $69.2 \pm 8.3$ \\
    None ($\priorD + \targetD$ data)  & $22.5 \pm 9.2$ & $32.3 \pm 9.8$ & $37.9 \pm 8.8$ & $29.2 \pm 8.3$ & $30.5$ & $28.4 \pm 10.9$ & $31.3 \pm 10.7$ & $13.9 \pm 5.5$ & $27.8 \pm 10.2$ & $25.4$ & $82.1 \pm 6.8$  \\
    \midrule
    Lang Reg. (ours) & $20.6 \pm 8.1$ & \boldmath{$57.3 \pm 8.1$} & $63.1 \pm 7.7$ & \boldmath{$32.5 \pm 6.3$} & $43.4$ & $54.0 \pm 7.2$ & \boldmath{$62.5 \pm 12.1$} & $76.0 \pm 8.7$ & $58.5 \pm 9.3$ & $62.8$ & $90.7 \pm 5.4$   \\
    Lang Dist. (ours) & $23.8 \pm 5.4$ & \boldmath{$57.3 \pm 10.6$} & $66.9 \pm 5.6$ & $27.9 \pm 10.8$ & $44.0$ & \boldmath{$65.5 \pm 13.1$} & $56.7 \pm 9.9$ & \boldmath{$78.6 \pm 5.1$} & $54.4 \pm 11.5$ & \boldmath{$63.8$} & $90.0 \pm 5.0$   \\
    Stage Classif. & \boldmath{$30.4 \pm 10.4$} & $52.7 \pm 6.0$ & \boldmath{$67.5 \pm 8.3$} & $27.9 \pm 7.1$ & \boldmath{$44.6$} & $63.1 \pm 9.9$ & $62.1 \pm 9.3$ & $55.4 \pm 8.5$ & \boldmath{$67.7 \pm 9.7$} & $62.1$ & \boldmath{$91.4 \pm 3.6$}   \\
    \midrule
    CLIP (frozen) & $1.7 \pm 0.4$ & $1.9 \pm 1.9$ & $3.8 \pm 2.5$ & $4.0 \pm 2.7$ & $2.9$ & $36.1 \pm 14.3$ & $39.9 \pm 8.9$ & $28.8 \pm 8.9$ & $48.4 \pm 11.9$ & $38.3$ & $75.6 \pm 7.7$  \\
    R3M (frozen) & $4.5 \pm 3.3$ & $9.0 \pm 4.8$ & $19.8 \pm 6.9$ & $15.4 \pm 5.4$ & $12.2$ & $49.4 \pm 11.6$ & $36.5 \pm 11.9$ & $47.0 \pm 14.1$ & $56.0 \pm 10.0$ & $47.2$ & $90.2 \pm 4.4$ \\
    \bottomrule
  \end{tabular}
  }
  \label{tab:sim}
\end{table*}
\subsubsection{\textbf{\texttt{sim2sim}}} For \texttt{sim2sim} experiments on this task suite, we define a new $\targetD$ simulated environment with differences from $\priorD$ as enumerated in Sec.~\ref{sec:exp:sim2sim-sim2real-diff}. Policies are trained with the 1600 $\priorD$ demonstrations and 100 target task $\targetDtargetT$ demonstrations.
% $\targetDpriorT$ consists of stacking trajectories but with a different one of the four objects than in $\targetDtargetT$.
%Note that $\targetD$, like $\priorD$, has the same four tasks that involve manipulating the four objects.  i thought this was kind of confusing

\subsubsection{\textbf{\texttt{sim2real}}} For \texttt{sim2real}, $\priorD$ remains the same as \texttt{sim2sim}. $\targetD$ is a real world environment in which the object is randomly placed on the left mat and the target task $\targetDtargetT$ is to move the object onto the right mat and open the gripper by the end of 20 timesteps.
% As a real world prior task $\targetDpriorT$, we collect 50 trajectories of moving a small paper box from the left to the right mat, while $\targetDtargetT$ involves moving a carrot. %Our method pretrains on the images from these 50 prior task trajectories without any other $\targetD$.

%%%%%%%%%%%%%%
%%% TASK 2 %%%
%%%%%%%%%%%%%%

\subsection{Task Suite 2: Multi-step Pick and Place}
Our second suite of tasks is longer-horizon. In simulation, the robot must first put an object in the pot, then grasp the pot by its handle and move it onto the stove. We categorize this as a 2-step pick-and-place task. We use the same four object-task mappings from Sec.~\ref{sec:exp:stack}. The object, pot, and stove locations are all randomized within a quadrant of the workspace. Since this task is longer horizon, we train on more data---$1,400$ trajectories per task in $\priorD$.

\subsubsection{\textbf{\texttt{sim2sim}}} Similar to the stacking task, in the \texttt{sim2sim} setting, we define a new $\targetD$ environment with differences from $\priorD$ enumerated in Sec.~\ref{sec:exp:sim2sim-sim2real-diff} and evaluate over the four tasks when given $100$ target-task $\targetDtargetT$ demonstrations.
% $\targetDpriorT$ contains a different task from the one in $\targetDtargetT$.

\subsubsection{\textbf{\texttt{sim2real}}} In the \texttt{sim2real} setup, $\priorD$ remains the same, while $\targetD$ is the real task of putting a carrot into a bowl, then putting the bowl onto a plate (see Fig.~\ref{fig:tasks}), and ending with the gripper open after 50 timesteps.
% $\targetDpriorT$ data consists of 50 trajectories, but involving a wooden block instead of a carrot.
In addition to success rate (Section~\ref{sec:exp:metric}), we measure average number of consecutive subtasks completed from the beginning, allowing partial credit if the robot only succeeds in the first step of placing the carrot in the bowl. However, if the robot does not finish the first step but finishes the second step, we do not count this as having completed any subtasks.

\subsection{Task Suite 3: Wrap Wire}
Our final suite of tasks involves wrapping a long deformable wire around a fixed object. In simulation $\priorD$, we approximate a wire with a chain of spheres connected with free joints, and the task is to wrap the chain around a fixed cylinder (see Fig.~\ref{fig:tasks}). A trajectory is successful if the first link of the chain has traveled $\geq \frac{5\pi}{3}$ radians ($5/6$ths of a full revolution) around the cylinder. Our simulation data consists of two tasks: wrapping counterclockwise and clockwise. The initial position of the end of the chain is randomized over a region to the left of the cylinder. $\priorD$ contains 400 trajectories per task. 

\subsubsection{\textbf{\texttt{sim2sim}}} For our $\targetD$ sim environment, we again apply the changes specified in Sec.~\ref{sec:exp:sim2sim-sim2real-diff}. We additionally swapped the spheres for capsules and changed the color and texture of the table, robot arm, and objects. This task has a wider sim2sim gap from additional visual and dynamics changes.
% We collect 100 trajectories for $\targetDpriorT$ that involves wrapping clockwise instead of counterclockwise in $\targetDtargetT$.

\subsubsection{\textbf{\texttt{sim2real}}} In our \texttt{sim2real} experiments, the target task $\targetDtargetT$ is to first grasp the plug, then wrap the cord around the base of a blender in the middle of the workspace, and finally put the plug down, similar to what one might do before putting the appliance away.
% For real world prior task $\targetDpriorT$ data, we collect 50 trajectories of a similar wrapping task, except the plug, cord, and blender are replaced by a wooden block, ethernet cable, and spool, respectively.
Like the sim environment, we define success if the following two conditions are met: (1) the plug travels $\geq \frac{5 \pi }{3}$ radians around the blender, and (2) the plug is placed and the gripper is open at the end of 50 timesteps. 

\subsection{Baselines}
To evaluate the effectiveness of \ouralgo{}, we consider two sets of baselines: non-pretrained baselines where the CNN is initialized from scratch, and baselines with pre-trained visual encoders.
For the non-pretrained baselines, we examine training with only $\targetD$ data, and training with both $\priorD$ and $\targetD$ data. This enables us to understand the benefits of our proposed training procedure.
\rev{In \texttt{sim2real}, we also compare to three popular prior sim2real approaches:}
\begin{itemize}
    \item \rev{\textbf{MMD}~\cite{tzeng2014deep}, which aims to minimize the distance between the mean embedding of all sim images and all real images of a batch to prevent the real images from being out-of-distribution relative to the sim images.}
    \item \rev{\textbf{Domain randomization}~\cite{tobin2017domain} of the colors, textures, and physics of the $\priorD$ environment.}
    \item \rev{Automatic Domain Randomization with Random Network Adversary (\textbf{ADR+RNA})~\cite{openai2019solving}, which keeps increasing/decreasing domain randomization bounds depending on the agent's performance, and also introduces a randomly initialized network for each trajectory to inject correlated noise into the agent's action conditioned on the state input.}
\end{itemize}
For the pretrained baselines, we consider two strong foundation models as the visual backbone, CLIP~\cite{radford2021learning} and R3M~\cite{nair2022r3m}, commonly used visual representations for robotics that are, like our approach, also shaped by language descriptions of images/videos, and add a trainable policy head composed of fully-connected layers sharing the same dimensions as all other methods in our results. %, as both are commonly used visual representations for robotics and were shaped by language descriptions of images/videos.

% Since our method uses extra language labels during pretraining that the baselines do not have access to, in all our experiments, we give all baselines an augmented $\targetDpriorT$ dataset that includes action-labeled demonstrations, in addition to the target task, $\targetDtargetT$. \textit{Note that our method is not given $\targetDpriorT$ action-labeled data}: it is trained only on $\targetDpriorT$ images with language labels during image-language pretraining (Sec.~\ref{sec:method:imglangpt}) but not during BC policy learning. Therefore, the baselines in a sense serve as upper bounds as they are given $\left| \targetDpriorT \right| = 50$ additional action-labeled demonstrations. In other words, during policy learning, the baselines train on action-labeled demonstrations from $\priorD \cup \targetDpriorT \cup \targetDtargetT$ while ours are only trained on $\priorD \cup \targetDtargetT$.
For each task in \texttt{sim2real}, we train and evaluate with 25, 50, or 100 $\targetDtargetT$ demonstrations.

\subsection{Our Method Variants and Ablations}
\label{exp-results:method-ablation}
In our evaluations, we compare language regression (Section~\ref{sec:method:lang-reg}) and language distance (Section~\ref{sec:method:lang-dist}), the two pretraining variants of our approach. We also ablate away the effects of language on our pretraining approach in a method called ``stage classification,'' where the pretraining task is to predict the stage index of an image (see Section~\ref{sec:method:langlabeling}) instead the language embedding or embedding distance.

\section{Experimental Results}
\label{sec:experimental-results}
Our results for \texttt{sim2sim} experiments are shown in Table~\ref{tab:sim}, and the results for \texttt{sim2real} are shown in Table~\ref{tab:real}. In both tables, the methods (rows) are grouped into non-pretrained baselines, our method variants and ablations, and pretrained SOTA baselines. In \texttt{sim2real}, we additionally include a group of three rows to show the performance of prior sim2real approaches.

\subsection{Experimental Questions and Analysis}
Across the three task suites in both \texttt{sim2real} and \texttt{sim2sim}, our method generally achieves the highest success rates.
% In particular, task suite 3 involves manipulating deformable objects, which is hard to simulate with high fidelity, widening the sim2real gap. We see lower success rates across the board but attribute this to small rigid plug (2cm $\times$ 5cm base) requiring a higher degree of grasping precision than the larger soft carrot in the previous tasks. Importantly, our method performs better than SOTA vision-language pretraining and non-pretrained baselines with wire wrap on 50 and 100 target task demonstrations on task suite 3, which demonstrates the effectiveness of \ouralgo{}.
% In general, we notice more pronounced differences between variants of our method and the  in sim2real, likely due to the larger domain gap.
% In sim2real, we evaluate on 3 levels of $\targetDtargetT$ data. In sim2sim we evaluate on 4 tasks for Task Suite 1 and 2. The predominant trends, however, do remain constant between sim2sim and sim2real.
To further analyze the effectiveness of our method, we pose and investigate the following experimental questions.

\textit{\textbf{What is the impact of our pretraining approach?}}
% Even without action-labeled $\targetDpriorT$ demonstrations that the baselines have access to, 
Our method nearly doubles the success rate of both non-pretrained baselines in most task suites in \texttt{sim2real} and \texttt{sim2sim}. This indicates that \ouralgo{} can bridge a wide sim2real gap. One factor that may allow our method to perform well is that image observations with similar language descriptions may also have similar action labels. In Appendix~\ref{appendix:lang-act-dist-sim}, we further investigate this hypothesis with an analysis of the action distributions between images, split by their language descriptions.

Between the non-pretrained baselines, training on $\priorD$ sim demonstrations in \texttt{sim2real} provides little benefit on stack object, increases average performance by $\approx20\%$ on multi-step pick-and-place, but decreases average performance by $\approx10\%$ on wrap wire. However, in \texttt{sim2sim}, it provides a $10$-$15\%$ increase on most tasks. This suggests that the sim2sim gap is small enough to benefit from using $\priorD$ even without pretraining, but that the sim2real gap is large enough for pretraining to be needed to leverage $\priorD$.
% The non-pretrained baseline using only $\targetD$ data $(\targetDpriorT \cup \targetDtargetT)$ and the second using $\targetD$ and $\priorD$ data, the $\priorD$ trajectories add little benefit without pretraining, given the large domain gap.

\textit{\rev{\textbf{How does our method compare to prior sim2real baselines?}}}
\rev{Our method outperforms all of the prior sim2real baselines we tested against (second row-group in Table~\ref{tab:real}), which collectively do relatively poorly in most settings, highlighting the difficulty of the sim2real problem in our setup.}

\rev{MMD averages the best performance across the three sim2real baselines and even achieves competitive performance on the easiest task of stacking an object. However, on the two other more difficult tasks, its performance does not scale well with more trajectories, which we suspect arises from stability issues in trying to push together the mean of all sim and real image embeddings in each batch. Domain randomization only exacerbates the sim2real gap since enabling all randomizations does not move the distribution of simulation trajectories closer to the real world trajectories due to the large visual dissimilarity between our simulation and real environments. ADR+RNA, which only randomizes the environment as much as possible without severely hurting the scripted policy performance, averages slightly better performance than domain randomization, perhaps because the data is less diverse and easier to fit a policy to than the data from full-scale domain randomization.}

\textit{{\textbf{How does our method compare to prior vision-language pretrained representations?}}}
In \texttt{sim2real}, our method outperforms both pretrained baselines across the board, including R3M, which is the strongest baseline on stack object and multi-step pick-and-place. When trained on increasing amounts of real-world data, both R3M and CLIP tend to plateau---CLIP performs no better than $40\%$ on any task, R3M has an apparent ceiling of $65\%$, while our method achieves up to $90\%$. This suggests that CLIP and R3M do not scale as well as our method when provided more data, despite being pretrained on internet-scale video and image data while our method was pretrained on images from just a few hundred sim and real trajectories.

In \texttt{sim2sim}, our method also outperforms R3M and CLIP across the board. Averaging the performance on stacking and multi-step pick-and-place, our method outperforms R3M by 15-30\% and CLIP by 25-40\%. On the wrap wire task, our method and R3M perform comparably, probably because the task is quite a bit easier for all methods in simulation.

\textit{\textbf{What is the effect of language in learning shared representations?}}
We ablate the effect of language on our pretraining as the ``stage classification'' row in Tables~\ref{tab:real} and~\ref{tab:sim}, as mentioned in Section~\ref{exp-results:method-ablation}. In \texttt{sim2sim}, we see similar performance in language regression pretraining and stage classification pretraining. However, in \texttt{sim2real}, where the domain gap is larger, we see language providing a measurable benefit in all task suites, especially in multi-step pick-and-place, perhaps because pretraining with language leverages similarities in language descriptions between the first and second steps of the pick-and-place task.
 % , while also differentiating the objects in each subtask.
% Conversely, the classification baseline does not benefit as much and performs worse on task 2, showing the advantage of language.

\textit{\textbf{How do our two image-language pretraining variants compare?}}
We compare our two pretraining variants introduced in Sections~\ref{sec:method:lang-reg} and~\ref{sec:method:lang-dist}, where language regression directly predicts language embeddings while language distance is encouraged to maintain pairwise distances based on BLEURT similarity scores. Again in \texttt{sim2sim}, there is no clear winner between the two, but in \texttt{sim2real}, language regression performs better on average. This suggests that when performing language pretraining for visual representations, the more constraining regression loss outperforms the less constraining distance-matching loss on \texttt{sim2real} performance.

\subsection{Additional Experimental Questions and Results}
Finally, we examine a few questions to better understand the performance of our method under slight changes to the data and problem setup.

\textit{\textbf{\rev{What is the effect of pretraining on image-language pairs where the language granularity is reduced?}}}
\rev{We evaluate the impact of reduced language granularity on \texttt{sim2real} performance. See Appendix~\ref{appendix:lang-granularity-results} for results.}

\textit{\textbf{How does our method perform if we cannot pretrain directly on image-language pairs from the target task?}}
There are scenarios in which we might not have access to the real-world target task $\targetDtargetT$ during the pretraining phase, as pretraining is often done without knowledge of the downstream task. To investigate this, we introduce a real-world prior task $\targetDpriorT$ that we pretrain on, and use real-world target task data $\targetDtargetT$ only during imitation learning. The advantage of this problem setup is that we can reuse the same $f_{cnn}$ for multiple downstream real-world target tasks as long as they are sufficiently similar to the real-world prior task. In this modified problem setup, our method still mostly outperforms all baselines, which demonstrates that our method does not overfit to the real-world task it sees during pretraining. See Appendix~\ref{appendix:sim2real-results-w-prior} for full results.

\textit{\textbf{Can our method be combined with prior large-scale vision-language pretrained networks?}}
We experiment with combining R3M and our method. Results are discussed in Appendix~\ref{appendix:r3m-ft-ours}. Similar to~\cite{zhai2021lit}, we find that finetuning R3M on our pretraining procedure does not yield improvements on downstream policy learning.

\textit{\textbf{What are some failure modes of our policy?}}
See our project website at \url{https://robin-lab.cs.utexas.edu/lang4sim2real/} for failure videos. In \textbf{stack object}, most failures arose from not closing the gripper enough before attempting to lift the object, since our gripper action dimension is analog (controls the distance between the gripper fingers) instead of binary (open/closed). In \textbf{multi-step pick-and-place}, most failures of our policy came from the robot not moving right after successfully completing the first pick-and-place operation. In \textbf{wrap wire}, most failures were from grasping imprecision, since the small plug allowed for only a $\leq3$ cm $xy$-plane error in the grasping point. Other common failures included prematurely opening the gripper and dropping the plug before the wrapping was finished, or moving the gripper sideways into the blender instead of properly circling around it.

% This demonstrates that directly predicting the language embeddings is more successful than distance learning. The higher amount of constraint on the learned representation, and the additional structure imposed by language embeddings is helps to preserve more information and \_\_\_. 

%should we have a section on sim2sim vs sim2real

%better explanation this is not clear
% via directly predicting language embeddings compare to distance learning, in which the CNN is encouraged to maintaining the BLEURT similarity between language and images?

\section{Conclusion} 
\label{sec:conclusion}
Vision-based policies struggle with distributional shift during sim2real transfer. To address this challenge, we introduced a low-data-regime visual pretraining approach that leverages language to bridge the sim2real visual gap with only 25-100 real-world trajectories with automatically generated language labels. We evaluate the effectiveness of our approach on multi-step long-horizon tasks and hard-to-simulate deformable objects. In the few-shot setting, our approach outperforms state-of-the-art vision-language foundation models and prior sim2real approaches across 3 task suites in both \texttt{sim2sim} and \texttt{sim2real}.

\section{Limitations and Future Work} 
\label{sec:limitations}

One of the main limitations of our work is that the learned representation may have limited generalizability compared to existing pretraining methods that leverage internet-scale data to enable a large degree of generalization. Our approach targets a specific distribution and domain of real-world tasks and operates in the low-data regime for both pretraining and policy learning, so it does not yield general-purpose visual representations that can be applied to a wide distribution of target tasks. Future work could investigate scaling our method to large-scale datasets to further reduce the number of real-world demonstrations needed for effective sim2real transfer.

% Our pretraining method results in learned representations that cannot be used entirely frozen, as it requires some additional finetuning on the last layer for policy data. This suggests that \ourmethod{} is just providing a good initialization for the final layers of the CNN and not necessarily providing a usable out-of-the-box frozen representation like other vision+language pretrained models for robotic control.

Our method also assumes that the template language descriptions used by the automatic labelling process describe similar aspects of images across the two domains, and may perform worse if the language between sim and real described images at extremely different levels of granularity. Furthermore, our approach relies on segmenting all trajectories of a task into stages of a certain granularity so that the associated template language is diverse enough to prevent the learned visual representation from mapping the entire input image distribution to a collapsed point. \rev{On contact-rich tasks involving continuous motions or complex object deformations, it may be harder to segment a trajectory and label these segments with language.}
% Along the same lines, it would be worth investigating the performance of our approach when trained on more diverse language descriptions, such as from a group of human annotators.

% There are situations where it's harder to get language labels on real trajectories than action labels.

% We assumed that similar language descriptions of images map to similar semantic action distributions. This holds in most of the cases in robot manipulation with a limited number of objects on the scene. Sometimes, there will be multimodality in the action distribution given a current image observation, such as if the robot must manipulate objects at different parts of the workspace. Our method is not tested in such scenarios.

Another avenue for future work involves exploring sim2real by combining existing pretraining approaches such as time-contrastive learning and masked image modeling in conjunction with the language-based pretraining we propose, as adding temporal or masked prediction terms to the objective may enable more fine-grained representations that complement the coarseness of language. 

% Additional investigations could focus on investigating multitask vs. transfer learning

\section{Acknowledgements}
We would like to thank members of the RobIn Lab at UT Austin for their valuable suggestions and help with debugging real robot issues. This research was supported by NSF NRI Grant IIS-1925082.

% "Acknowledgments to people or funding agencies should not appear in the submission." (https://roboticsconference.org/information/authorinfo/)
% \section*{Acknowledgments}

%% Use plainnat to work nicely with natbib. 

\bibliographystyle{plainnat}
\bibliography{references}
\newpage
\section*{Appendix}
\subsection{Scripted Policy for Real-World Data Collection}
\label{appendix:scripted-policies}

\begin{minipage}[htb]{\linewidth - .1in}
    \begin{algorithm}[H]
        \footnotesize
        \caption{Scripted Wrap Wire}
        % \vspace{-0.2in}
        \label{alg:scripted-wrap-wire}
        \newcommand{\MYCOMMENT}[1]{~\textcolor{gray}{// #1}}
            % \textbf{Input:} $\mathcal{D}$ buffer
        \begin{algorithmic}[1]
            \State centerPos $\leftarrow$ blender center position
            \State placeAttempted $\leftarrow$ False
            \State targetDistToCenter $\leftarrow$ 0.15
            \State numTimesteps $\leftarrow$ 45
            \State direction $\leftarrow$ true if clockwise, false if counterclockwise
        
            \For {t \textbf{in} [0, numTimesteps)}
                \State wirePos $\leftarrow$ position of the graspable part of the wire
                \State eePos $\leftarrow$ end effector position
                \State pickPosDist $\leftarrow \| \text{eePos} - \text{wirePos} \|_2$
                \State done $\leftarrow$ is wrapped $> \frac{11\pi}{6}$ from the start to end of wire around centerPos in direction
                \If {placeAttempted}
                    \State action $\leftarrow 0$
                    \ElsIf {object not grasped \textbf{AND} pickPosDist $>$ distThresh}
                    \State \MYCOMMENT{Move toward wire}
                    \State action $\leftarrow$ wirePos $-$ eePos
                \ElsIf {object not grasped}
                    \State \MYCOMMENT{gripper is very close to wire}
                    \State action $\leftarrow$ pickPos $-$ eePos
                    \State close gripper \MYCOMMENT{Object is in gripper}
                \ElsIf {wire not lifted}
                    \State action $\leftarrow$ [0, 0, 1] \MYCOMMENT{Move up}
                \ElsIf {not done}
                    \State relPos $\leftarrow \text{eePos} - \text{centerPos}$
                    \State distToCenter $\leftarrow \| \text{relPos}\|_2$
                    \State normRelPos $\leftarrow (\text{relPos} / \text{distToCenter}) * \text{targetDistToCenter}$
                    \State actionMaintainDistance $\leftarrow \text{relPos} * (\text{targetDistToCenter} - \text{distToCenter})$ \MYCOMMENT{move toward/away from center}
                    \State actionMoveTangent $\leftarrow [-\text{normRelPos}[1], \text{normRelPos}[0], 0.0]$ \MYCOMMENT{Move tangent to the blender}
                    \If {direction}
                        \State actionMoveTangent $\leftarrow \text{actionMoveTangent} * -1$
                    \EndIf
                    \State action $\leftarrow \text{actionMaintainDistance} + \text{actionMoveTangent}$
                \Else
                    \State action $\leftarrow$ open gripper \MYCOMMENT{Drop wire}
                    \State placeAttempted $\leftarrow$ True
                \EndIf         
            \EndFor            
        \end{algorithmic}
       
    \end{algorithm}
\end{minipage}

\begin{minipage}[htb]{\linewidth - .1in}
    \begin{algorithm}[H]
      	\footnotesize
      	\caption{Scripted Pick and Place Function}
        % \vspace{-0.2in}
      	\label{alg:scripted-pick-place}
        \newcommand{\MYCOMMENT}[1]{~\textcolor{gray}{// #1}}
        % \textbf{Input:} $\mathcal{D}$ buffer
      	\begin{algorithmic}
      	    
    \Function{PickPlace}{pickPos, dropPos, distThresh, placeAttempted}

        \State eePos $\leftarrow$ end effector position
        \State dropPosDist $\leftarrow \| \text{eePos} - \text{dropPos} \|_2$
        \State pickPosDist $\leftarrow \| \text{eePos} - \text{pickPos} \|_2$
        \If {placeAttempted}
            \State action $\leftarrow 0$
        \ElsIf{object not grasped \textbf{AND} pickPosDist $>$ distThresh}
            \State \MYCOMMENT{Move toward target object}
            \State action $\leftarrow$ pickPos $-$ eePos
        \ElsIf {object not grasped}
            \State \MYCOMMENT{gripper is very close to object}
            \State action $\leftarrow$ (pickPos $-$ eePos, close gripper) \MYCOMMENT{Object is in gripper}
        \ElsIf {object not lifted}
            \State \MYCOMMENT{Move gripper upward to avoid hitting other objects/containers}
            \State action $\leftarrow$ [0, 0, 1]
        \ElsIf {dropPosDist $>$ distThresh}
            \State \MYCOMMENT{Move toward target container}
            \State action $\leftarrow$ dropPos $-$ eePos
        \Else
            \State action $\leftarrow$ open gripper \MYCOMMENT{Object falls into container}
            \State placeAttempted $\leftarrow$ True
        \EndIf
        \State noise $\sim \mathcal N(0, 0.1)$
        \State action $\leftarrow$ action + noise\\       
        \Return action, placeAttempted
        \EndFunction

        % \STATE pickPos $\leftarrow$ target object position
        % \STATE dropPos $\leftarrow$ target container position
        % \STATE distThresh $\leftarrow 0.02$
        % \STATE numTimesteps $\leftarrow 30$
        % \STATE placeAttempted $\leftarrow$ False
        % \FOR {t \textbf{in} [0, numTimesteps)}
        %     \STATE eePos $\leftarrow$ end effector position
        %     \STATE dropPosDist $\leftarrow \| \text{eePos} - \text{dropPos} \|_2$
        %     \STATE pickPosDist $\leftarrow \| \text{eePos} - \text{pickPos} \|_2$
        %     \IF {placeAttempted}
        %         \STATE action $\leftarrow 0$
        %     \ELSIF {object not grasped \textbf{AND} pickPosDist $>$ distThresh}
        %         \STATE \MYCOMMENT{Move toward target object}
        %         \STATE action $\leftarrow$ pickPos $-$ eePos
        %     \ELSIF {object not grasped}
        %         \STATE \MYCOMMENT{gripper is very close to object}
        %         \STATE action $\leftarrow$ pickPos $-$ eePos
        %         \STATE close gripper \MYCOMMENT{Object is in gripper}
        %     \ELSIF {object not lifted}
        %         \STATE \MYCOMMENT{Move gripper upward to avoid hitting other objects/containers}
        %         \STATE action $\leftarrow$ [0, 0, 1]
        %     \ELSIF {dropPosDist $>$ distThresh}
        %         \STATE \MYCOMMENT{Move toward target container}
        %         \STATE action $\leftarrow$ dropPos $-$ eePos
        %     \ELSE
        %         \STATE open gripper \MYCOMMENT{Object falls into container}
        %         \STATE placeAttempted $\leftarrow$ True
        %     \ENDIF
        %     \STATE noise $\sim \mathcal N(0, 0.1)$
        %     \STATE action $\leftarrow$ action + noise
        %     \STATE $ s' \leftarrow$ env.step(action)
            
        % \ENDFOR
        
      	\end{algorithmic}
    \end{algorithm}
\end{minipage}

\begin{minipage}[htb]{\linewidth-.1in}
    \begin{algorithm}[H]
      	\footnotesize
      	\caption{Stack Object}
        % \vspace{-0.2in}
      	\label{alg:stack-obj}
        \newcommand{\MYCOMMENT}[1]{~\textcolor{gray}{// #1}}
        % \textbf{Input:} $\mathcal{D}$ buffer
      	\begin{algorithmic}[1]
                \State pickPos $\leftarrow$ target object position
                \State dropPos $\leftarrow$ target container position
                \State numTimesteps $\leftarrow 18$
                \State distThresh $\leftarrow 0.02$
                \State placeAttempted $\leftarrow$ False
                \For {t \textbf{in} [0, numTimesteps)}
                     \State action, placeAttempted $\leftarrow$ \Call{PickPlace}{pickPos, dropPos, distThresh, placeAttempted}
                    \State $ s' \leftarrow$ env.step(action)
                \EndFor
       \end{algorithmic}
    \end{algorithm}
\end{minipage}

\begin{minipage}[htb]{\linewidth-.1in}
    \begin{algorithm}[H]
      	\footnotesize
      	\caption{Scripted 2-step Pick and Place}
        % \vspace{-0.2in}
      	\label{alg:scripted-2-step-pick-place}
        \newcommand{\MYCOMMENT}[1]{~\textcolor{gray}{// #1}}
        % \textbf{Input:} $\mathcal{D}$ buffer
      	\begin{algorithmic}[1]
                \State pickPos $\leftarrow$ [object position, first container position]
                \State dropPos $\leftarrow$ [first container position, second container position]
                \State numTimesteps $\leftarrow 45$
                \State distThresh $\leftarrow 0.02$
                \State placeAttempted $\leftarrow$ [False, False]
                \State  $s_i \leftarrow$ 0 \MYCOMMENT{step index (starts at 0, and increments to 1 when first pick-place step is complete) }
                \State stepCompleted $\leftarrow$ [False, False]

                \For {t \textbf{in} [0, numTimesteps)}
                    \State action, placeAttempted[$s_i$] $\leftarrow$ \Call{PickPlace}{pickPos[$s_i$], dropPos[$s_i$], distThresh, placeAttempted[$s_i$]}
                    \If {stepIsSuccessful($s_i$) \textbf{AND} \textbf{not} stepsCompleted[$s_i$]}
                        \State stepsCompleted[$s_i$] $\leftarrow$ True
                        \State $s_i \leftarrow 1$
                    \EndIf
                    \State $ s' \leftarrow$ env.step(action)
                    
                \EndFor
       \end{algorithmic}
    \end{algorithm}
\end{minipage}

\subsection{Detailed Policy Network Architecture \& Hyperparameters}
\begin{figure*}[h]
\includegraphics[width=2\columnwidth]{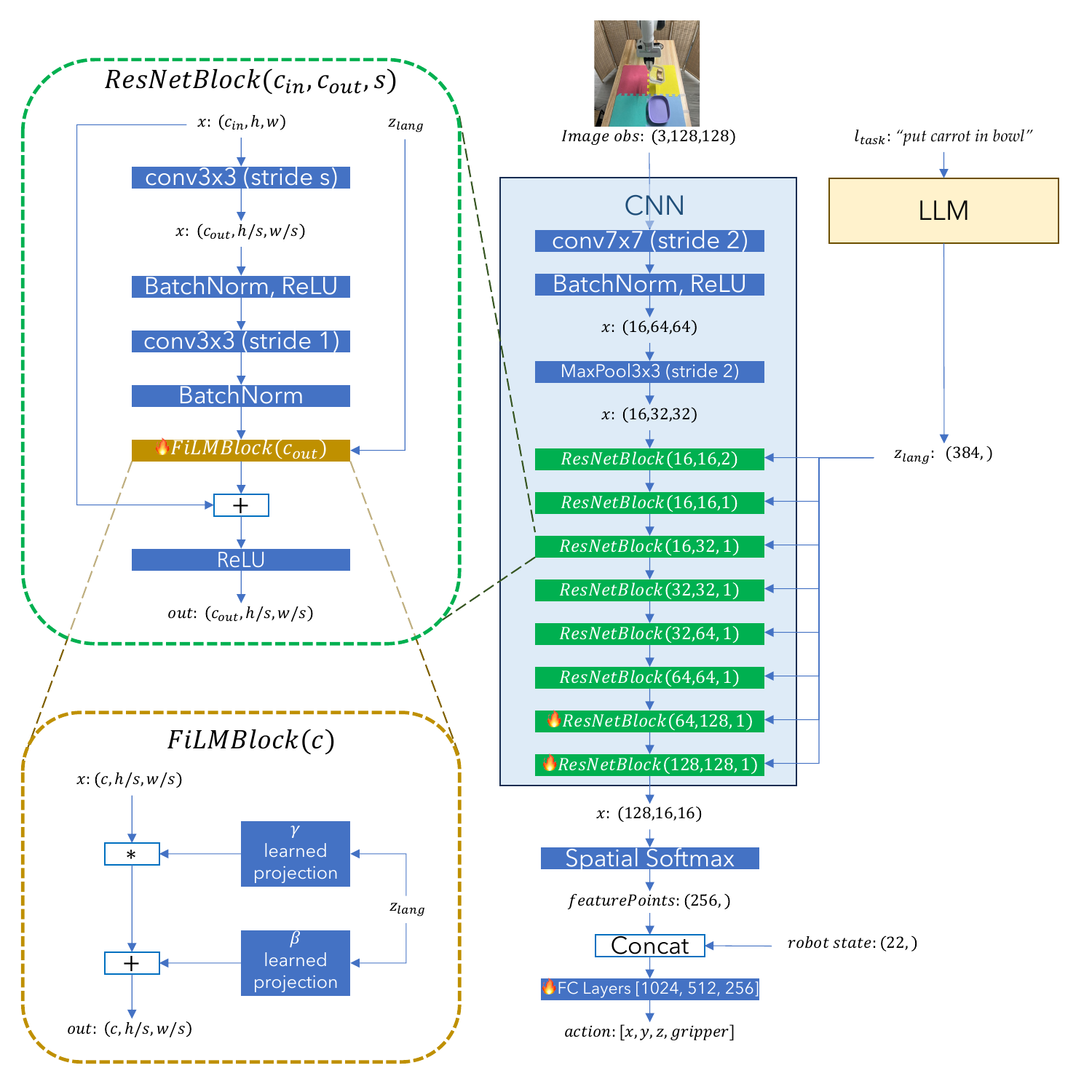}
\caption{Detailed Policy Network Architecture. Fire denotes layers trained during policy learning. The early CNN modules are kept frozen to maintain the intermediate representations learned from the pretraining phase.}
\label{fig:detailed-net-arch}
\end{figure*}

 For the policy backbone, we use a ResNet-18 architecture but made changes to the strides and number of channels to adapt the network to our $128 \times 128 \times 3$ image size. Hyperparameters are shown in Table~\ref{tab:policy-hparams}. A detailed layer-by-layer architecture figure of our policy is shown in Figure~\ref{fig:detailed-net-arch}. During policy training, only the last CNN layer, FiLM blocks, and policy head (FC layers) are finetuned, while all other layers are kept frozen.

\subsection{Does Language Similarity Imply Action Distribution Similarity?}
\label{appendix:lang-act-dist-sim}
\begin{figure*}[htb!]
\includegraphics[width=2\columnwidth]{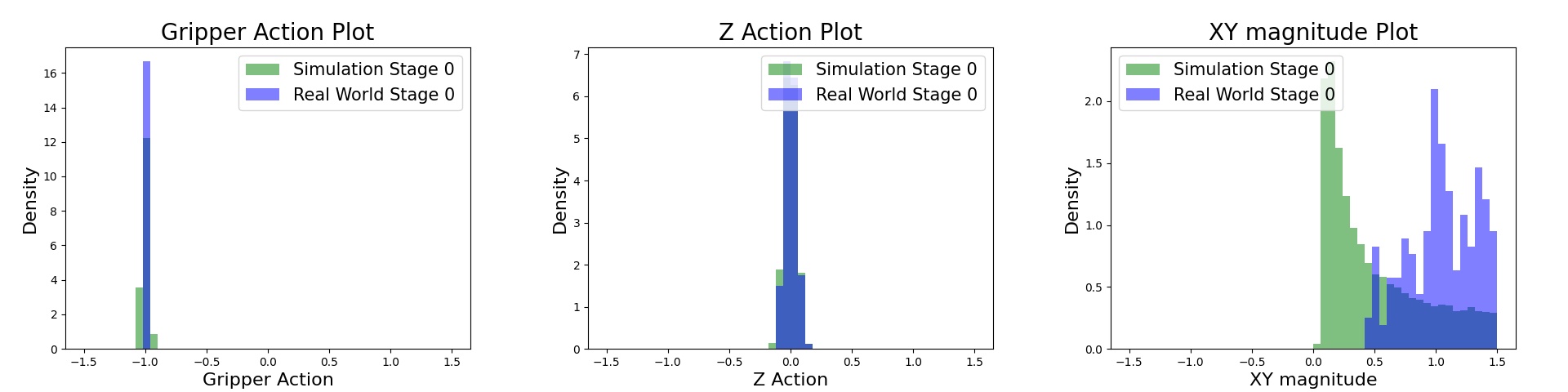}

\includegraphics[width=2\columnwidth]{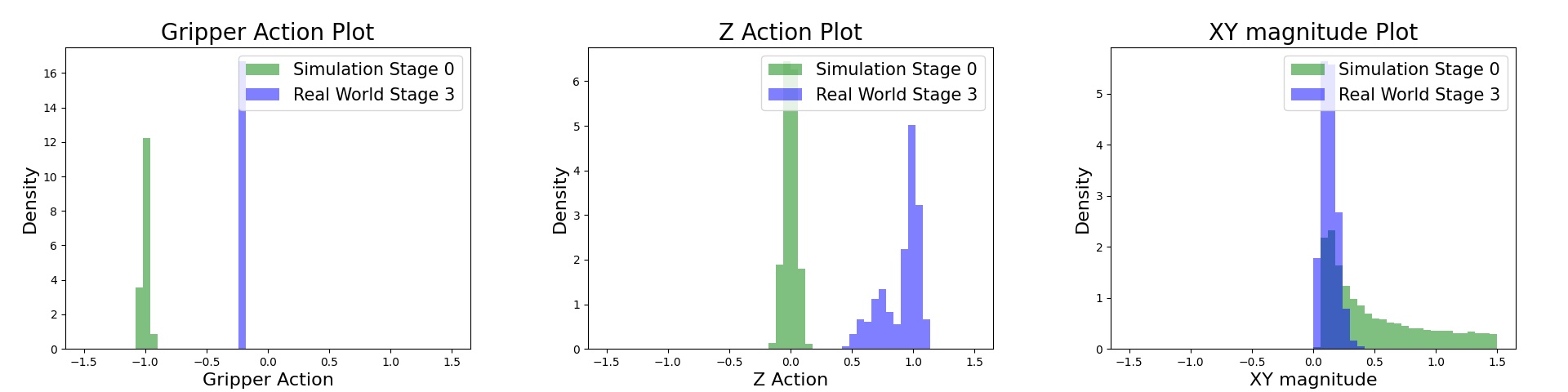}
\caption{These plots show the action distribution of demonstrations across both sim and real, broken down by each component of the action: $xy$-action magnitude, $z$-axis actions, and gripper actions. The first row shows simulation (green) and real world (blue) action distributions for images described by similar language. The second row shows the same distribution of simulation actions (green) as in the first row, but compared with real-world action distributions from images labeled with very different language from the sim actions (blue).
% , the simulation distribution still drawn from stage 0, while the real action distributions are now drawn from stage 3.
Notably, the action distributions are generally similar for images with similar language (first row), and different for images with different language (second row). This suggests that pretraining our CNN on language embedding prediction benefits downstream policy learning because it allows the domain-invariant learned representations to tap into similar action distributions for completing a task.}
\label{fig:action-dist}
\end{figure*}

%might need to reformat this, maybe 2x3 so you can see them easier
We hypothesize that one of the ways language is an effective bridge for sim2real transfer is that the sim and real action distributions of the demonstrations are similar when the image observations have similar language descriptions. Figure \ref{fig:action-dist} shows the action distribution similarities between sim and real when the language descriptions are similar (top row), and when the language descriptions are different (bottom row). Each column represents a component of the action distribution. We plot three components: $z$-axis actions, $xy$-magnitude (which is the $\ell 2$ norm of the $(x, y)$ action dimensions), and the gripper dimension. We observe that action distributions are indeed more similar for images described by similar language than for images described by different language.

\begin{table*}[h]
    \begin{center}
    \caption{Language Description Templates of Image Observations}
    \label{tab:lang-templates}
    \resizebox{\textwidth}{!}{
    \begin{tabular}{ll}
    \hline
     \rowcolor{LightCyan}
     \textbf{Task} & \textbf{Template String}\\
    \hline
     Pick and Place & gripper open, reaching for \textit{\{objName\}}, out of \textit{\{contName\}}\\
      & gripper open, moving down over \textit{\{objName\}}, out of \textit{\{contName\}}\\
      & gripper closing, with \textit{\{objName\}}, out of \textit{\{contName\}}\\
      & gripper closed, moving up with \textit{\{objName\}}, out of \textit{\{contName\}}\\
      & gripper closed, moving sideways with \textit{\{objName\}}, out of \textit{\{contName\}}\\
      & gripper closed, with \textit{\{objName\}}, above \textit{\{contName\}}\\
      & gripper open, dropped \textit{\{objName\}}, in \textit{\{contName\}} \\
      \hline
      Wrap Wire & gripper open, reaching for \textit{\{graspObjName\}}\\
        & gripper open, moving down over \textit{\{graspObjName\}}\\
        & gripper closing around \textit{\{graspObjName\}}\\
        & gripper closed, moving up with \textit{\{graspObjName\}}\\
        & \textit{\{direction\}} left\\
        & \textit{\{direction\}} front\\
        & \textit{\{direction\}} right\\
        & \textit{\{direction\}} back\\

        & gripper open, above \textit{\{graspObjName\}} with \textit{\{flexWraparoundObjName\}} fully wrapped\\    & gripper open, above \textit{\{graspObjName\}} with \textit{\{flexWraparoundObjName\}} fully unwrapped\\
        \hline 
        \rowcolor{LightCyan}
        \textbf{Variable} & \textbf{Possible Values}\\
        \hline 
        \textit{objName} & milk, bread, can, cereal, pot, carrot, bowl, bridge \\
        \textit{contName} & coaster, pot, stove, bowl, plate\\
        \hline
        \textit{flexWraparoundObjName} & beads, cord, ethernet cable\\
        \textit{graspObjName} & last bead, white plug, bridge\\
        \textit{direction} & clockwise, counterclockwise \\
    \hline
     \end{tabular}
     }

     \end{center}

\end{table*}

\subsection{Task and Data Details}
\begin{figure*}[h!]
\includegraphics[width=2\columnwidth]{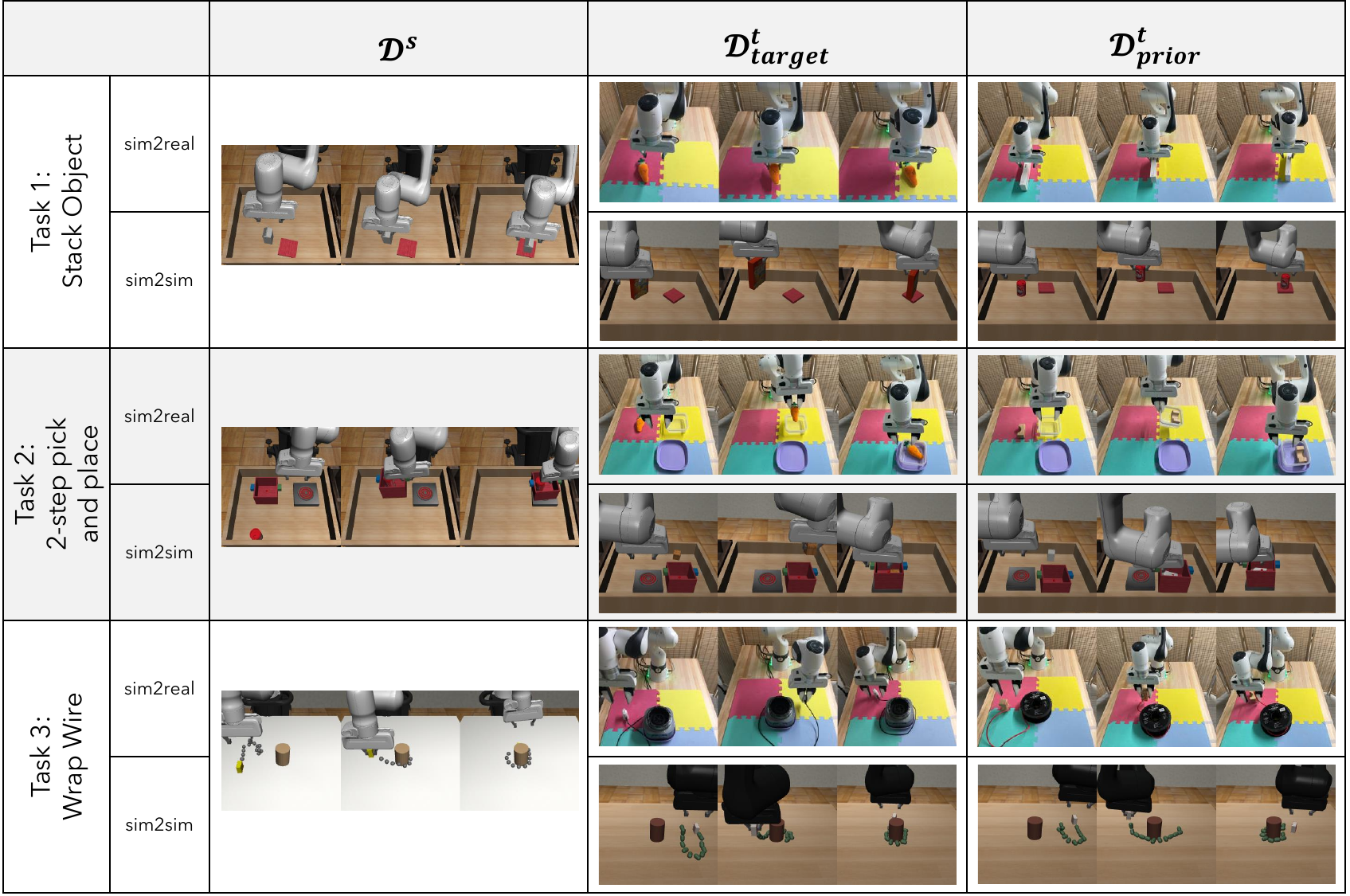}
\caption{This table builds on Figure~\ref{fig:tasks} and depicts the 3 datasets for each task with filmstrips. The rows show the three task suites while each column represents one of the three datasets we use during pretraining or policy learning. Our main results in Tables~\ref{tab:real} and \ref{tab:sim} use $\priorD \cup \targetDtargetT$ for pretraining and policy learning, whereas our results in Table~\ref{tab:real-w-prior} use $\priorD \cup \targetDpriorT$ for pretraining and $\priorD \cup \targetDtargetT$ for policy learning. This table shows the visual differences between sim and real, as well as the task in $\targetDpriorT$ versus $\targetDtargetT$.}
\label{fig:film-strips}
\end{figure*}

Figure \ref{fig:film-strips} provides film strips of trajectories from the source domain data $\priorD$, target domain prior task data $\targetDpriorT$, and target domain target task data $\targetDtargetT$, for each of the three task suites.

\begin{table*}[htbp]
  \centering
  \caption{\texttt{sim2real}: \rev{Performance in $\priorD \cup \targetDtargetT \cup \targetDpriorT$ setting by Number of Target Task Demonstrations}}
  \resizebox{\textwidth}{!}{
  \begin{tabular}{cccccccccccccccc}
    \toprule
    Method & \multicolumn{3}{c}{Action-labeled Data} &  \multicolumn{3}{c}{Stack Object} & \multicolumn{6}{c}{Multi-step Pick and Place} & \multicolumn{3}{c}{Wrap Wire}\\
    \cmidrule(r{4pt}){2-4} \cmidrule(l){5-7} \cmidrule(l){8-13} \cmidrule(l){14-16}
    & Sim & \multicolumn{2}{c}{Real} & \multicolumn{3}{c}{Success Rate \rev{(\%)}} & \multicolumn{3}{c}{Success Rate \rev{(\%)}} & \multicolumn{3}{c}{Subtasks Completed} & \multicolumn{3}{c}{Success Rate \rev{(\%)}} \\
    \cmidrule(l){5-7}\cmidrule(l){8-10}\cmidrule(l){11-13} \cmidrule(l){14-16}
    &  $\priorD$ &  $\targetDtargetT$ &  $\targetDpriorT$& 25 & 50 & 100 & 25 & 50 & 100 & 25 & 50 & 100 & 25 & 50 & 100 \\
    \midrule 
    
    No Pretrain ($\targetD$ data only) & -- & \ding{51} & \ding{51} & 45 & 30 & 65 & 40 & 20 & 30 & 1.15 & 0.9 & 1.15 &  25 & 45 & 35\\
     No Pretrain ($\priorD + \targetD$ data) & \ding{51} & \ding{51} & \ding{51} & 20 & 55 & 25 & 45 & 30 & 50 & 1.25 & 1.2 & 1.4 & 15 & 30 & 30 \\
    \midrule
    \rev{MMD} & \rev{\ding{51}} & \rev{\ding{51}} & \rev{\ding{51}} & \rev{35} & \rev{30} & \rev{40} & \rev{70} & \rev{45} & \rev{35} & \rev{\textbf{1.65}} & \rev{1.25} & \rev{1.2} & \rev{15} & \rev{0} & \rev{20} \\
    \rev{Domain Random.} & \rev{\ding{51}} & \rev{\ding{51}} & \rev{\ding{51}}  & \rev{25} & \rev{45} & \rev{60} & \rev{15} & \rev{15} & \rev{20} & \rev{0.9} & \rev{0.55} & \rev{0.85} & \rev{0} & \rev{5} & \rev{5} \\
    \rev{ADR+RNA} & \rev{\ding{51}} & \rev{\ding{51}} & \rev{\ding{51}} & \rev{15} & \rev{10} & \rev{20} & \rev{50} & \rev{5} & \rev{50} & \rev{1.35} & \rev{0.7} & \rev{1.25} & \rev{15} & \rev{10} & \rev{20} \\
    
    \midrule
    Lang Reg. (ours) & \ding{51} & \ding{51} & -- & 50 & 55 & \textbf{85} & 55 & \textbf{80} & \textbf{95} & 1.2 & \textbf{1.8} & \textbf{1.95} & 25 & \textbf{50} & 55\\
    Lang Dist. (ours) & \ding{51} & \ding{51} & -- & 30 & 65 & 70 & 25 & 50 & 65 & 0.95 & 1.4 & 1.5 & 15 & 25 & 60  \\
    Stage Classif. & \ding{51} & \ding{51} & -- & 70 & 60 & 70 & 20 & 60 & 85 & 0.9 & 1.5 & 1.8 & 15 & 20 & \textbf{70} \\
    \midrule
    CLIP (frozen) & \ding{51} & \ding{51} & \ding{51} & 30 & 25 & 35 & 25 & 45 & 35 & 0.55 & 0.95 & 0.95 & \textbf{35} & 40 & 45\\ 
    R3M (frozen) & \ding{51} & \ding{51} & \ding{51} & \textbf{80} & \textbf{70} & 80 & \textbf{75} & 75 & 85 & 1.6 & 1.55 & 1.75 & 30 & 25 & 20\\
    \bottomrule
  \end{tabular}
  }
  \label{tab:real-w-prior}
\end{table*}

\begin{table}[h]
    \begin{center}
    \caption{\textbf{Policy $\pi$ hyperparameters.}}
    \begin{tabular}{lr}
    \hline
     Attribute & Value\\
    \hline
     Input Height & 128 \\
     Input Width & 128 \\
     Input Channels & 3 \\
     Number of Kernels & [16, 32, 64, 128] \\
     Kernel Sizes & [7, 3, 3, 3, 3] \\
     Conv Strides & [2, 2, 1, 1, 1] \\
     Maxpool Stride & 2 \\
     Fully Connected Layers & [1024, 512, 256] \\
     Hidden Activations & ReLU \\
     FiLM input size & 384 \\
     FiLM hidden layers & 0 \\
     Spatial Softmax Temperature & 1.0 \\
     Learning Rate & $3 \times 10^{-4}$ \\
     Policy Action Distribution & Multivariate Isotropic Gaussian $\mathcal{N}(\mu, \sigma)$ \\
     Policy Outputs & $(\mu, \sigma)$ \\
     Image Augmentation & Random Crops \\
     Image Augmentation Padding & 4 \\ 
    \hline
    \label{tab:policy-hparams}
    \end{tabular}
    \end{center}
\end{table}

\subsection{Training Hyperparameters}

Table~\ref{tab:training-hparams} shows our BC training hyperparameters.

In each training iteration, we sample 4 random tasks from our training buffer and get 57 samples per task, for a total batch size of 228.
 
\begin{table}[H]
    \begin{center}
    \caption{\textbf{Imitation learning hyperparameters.}}
    \begin{tabular}{lr}
    \hline
     Attribute & Value\\
    \hline
     Number of Tasks per Batch & 4 \\
     Batch Size per Task & 57 \\
     Learning Rate & $3 \times 10^{-4}$ \\
    \hline
    \label{tab:training-hparams}
    \end{tabular}
    \end{center}
\end{table}

\subsection{Sim2Sim and Sim2Real Differences}
\label{appendix:sim2sim-sim2real-diff}

In our \texttt{sim2sim} experiments, $\priorD$ and $\targetD$ are both sim environments with the following differences:
\begin{enumerate}
    \item \textbf{Camera point-of-view}: $\priorD$ image observations are third person (looking toward the robot), and $\targetD$ image observations are first person (over the shoulder), a large change of viewing angle.
    \item \textbf{Friction and Damping}: Joint friction and damping coefficients are $5 \times$ and $50\times$ higher in $\targetD$ than $\priorD$, which significantly changes the dynamics.
\end{enumerate}

In our \texttt{sim2real} experiments, $\priorD$ in sim and $\targetD$ in real have the following differences:
\begin{enumerate}
    \item \textbf{Control frequency}: The simulated $\priorD$ policy runs at 50Hz while the real world $\targetD$ policy runs at 2Hz.
    \item \textbf{Objects}: The objects on the scene in each task differ between simulation and real data, except the robot itself.
    \item \textbf{Visual Observation}: Backgrounds and camera angles are markedly different between the two domains.
    \item \textbf{Initial positions}: The initial object and robot positions are different across sim and real.
    % \item \textbf{Action magnitude}: Real robot x, y, and z action are restricted so each are $\le 1$, and higher values are scaled before being executed. 
\end{enumerate}

\subsection{Labeling Image Observations with Language}
\subsubsection{\textbf{\rev{Language labeling during Scripted Policy}}}
\label{appendix:lang-labeling}
We automatically label image observations with language descriptions during the scripted policy data collection process. Each image is assigned a stage number \rev{based on the if-case of the scripted policy, which corresponds to a semantic positional arrangement between the gripper and the relevant objects on the scene.} Stage numbers map 1-to-1 to the template language strings shown in Table~\ref{tab:lang-templates}.

For example, for the pick-and-place/stack object task, we define 7 stages and 7 corresponding language string templates, where the first stage is when the gripper moves toward a point above the object, the second stage is when the gripper moves downward toward the object, and so on. For the 2-step pick-and-place task, we use 14 stages---2 consecutive lists of the 7 individual pick place string templates, where the object and container variables of each template are filled in with the proper names.

Though our approach to labeling image observations with language was done during demonstration collection, we emphasize that images can be automatically labeled with language in hindsight after demonstrations are collected.
For instance, one can run an object detector on the images to estimate the position of the gripper in relation to the scene objects. This information can be used to determine what stage in a pick-and-place trajectory an image observation falls into.

\subsubsection{\textbf{\rev{Alternative Approach: Language labeling with off-the-shelf VLMs}}}
\label{appendix:lang-labeling-vlm}
\rev{To relax the requirement that our automated language labeling process must occur synchronously with a scripted policy collecting demonstrations, we implemented an alternative approach that is decoupled from the demonstration collection process. First, we use an off-the-shelf open-vocabulary object detector model, GroundingDINO~\cite{liu2023grounding}, to output bounding boxes for the relevant objects on the scene. No finetuning of GroundingDINO is required.
Second, we train a CNN-based gripper state predictor to predict the gripper position $(x, y, z)$ as well as whether the gripper is opened or closed in a given image. This network is trained on previously collected $($image, gripper position, gripper opened/closed$)$ data from 100 trajectories, and takes one minute to train on a single A5000 GPU.}
\rev{Using these two models, we can get the gripper state and position relative to the objects, enabling us to predict a stage number that corresponds fairly closely with the actual stage number as outputted by our scripted policy. Finally, we verified that training our method on VLM-derived language annotations does not degrade performance. We performed image-language pretraining with language labels from either labeling method and tested on 2-step pick-and-place with 100 real-world trajectories. Both methods achieve 90\% success rate averaged over 2 seeds.}

\subsection{\rev{Impact of Language Granularity on Performance}}
\label{appendix:lang-granularity-results}
\rev{To examine the impact of decreasing language granularity on \texttt{sim2real} performance, we segment the trajectories into fewer and fewer stages, until the extreme case where the entire trajectory has only a single stage, which means that all images across all trajectories of a task have the same exact language description embedding. The language descriptions we use for each stage, for varying numbers of stages per task, are displayed in Tables~\ref{tab:language-2pp} (2-step pick-and-place) and \ref{tab:language-ww} (wire wrap). 
}

\rev{Results are shown in Table~\ref{tab:real-stages}. The trend is noisy, but in general, decreasing language granularity hurts performance slightly. Still, our method is robust to lower granularity, which matches our hypothesis that our pretraining approach provides significant performance gains simply by pushing sim and real images into a similar embedding distribution even if the language granularity is extremely coarse.}

\subsection{\rev{\texttt{sim2real} results with no pretraining on $\targetDtargetT$}}
\label{appendix:sim2real-results-w-prior}
\rev{In Tables~\ref{tab:real} and \ref{tab:sim}, we presented results in a setting where we both pretrained and did policy learning on two datasets, $\priorD$ and $\targetDtargetT$. Sometimes it is unrealistic to assume that during pretraining, we have access to the downstream target task we are ultimately interested in. In such scenarios, it may be more realistic to assume we instead have real-world data for a prior task, $\targetDpriorT$. Thus, in this setting, we experiment with pretraining on $\priorD \cup \targetDpriorT$ and training our policy on $\priorD \cup \targetDtargetT$.}

Our method uses extra language labels during pretraining that the baselines do not have access to. While these language labels can be acquired at scale, to compensate for this data advantage, we decided to give all baselines an augmented $\targetDpriorT$ dataset that includes action-labeled demonstrations, in addition to the target task, $\targetDtargetT$. \textit{Note that our method is not given $\targetDpriorT$ action-labeled data}: it is trained only on $\targetDpriorT$ images with language labels during image-language pretraining (Sec.~\ref{sec:method:imglangpt}) but not during BC policy learning. Therefore, the baselines in a sense serve as upper bounds as they are given $\left| \targetDpriorT \right| = 50$ additional action-labeled demonstrations. In other words, during policy learning, the baselines train on action-labeled demonstrations from $\priorD \cup \targetDpriorT \cup \targetDtargetT$ while ours are only trained on $\priorD \cup \targetDtargetT$. Results are shown in Table~\ref{tab:real-w-prior}.

\textit{\textbf{How different are $\targetDpriorT$ and $\targetDtargetT$?}} In \texttt{sim2sim} and \texttt{sim2real} for stack object and 2-step pick-and-place, the robot interacts with different objects in the two real-world tasks. Instead of a carrot as in $\targetDtargetT$, in $\targetDpriorT$, the robot interacts with a paper box for the stack object task suite and a rigid toy wooden block for 2-step pick-and-place.

In \texttt{sim2sim} on wire wrap, $\targetDpriorT$ contains data of the beads being wrapped clockwise, instead of counterclockwise in $\targetDtargetT$. In \texttt{sim2real} for wire wrap, the plug, cord, and blender in $\targetDtargetT$ are replaced by a wooden block, ethernet cable, and spool, respectively, in $\targetDpriorT$ data.
The differences between $\targetDpriorT$ and $\targetDtargetT$ can be visually examined in Figure~\ref{fig:film-strips}.

\textit{\textbf{What trends are different between Table~\ref{tab:real-w-prior} (with $\targetDpriorT$) and Table~\ref{tab:real} (without $\targetDpriorT$)?}} Most of the trends are similar. Re-examining our main experimental questions, we see that our method still nearly doubles the success rate of both non-pretrained baselines, outperforms all three prior sim2real baselines, and that using language regression is important to achieve the most gains from pretraining (language regression outperforms stage classification and language distance, on average). However, in this new problem setting in \texttt{sim2real}, R3M outperforms our method in the lowest data regime with 25 target task demonstrations, perhaps because of the additional 50 $\targetDpriorT$ demonstrations that our method does not train on. However, on 50 and 100 trajectories for the longer-horizon multi-step pick and place task, our method achieves higher sim2real performance than the best of either pretrained baseline.

\begin{table*}[t]
  \centering
  \caption{\rev{\texttt{sim2real}: Performance with Varying Language Granularity}}
  % \resizebox{\textwidth/2}{!}{
  % \begin{small}
  \begin{tabular}{cccccccccc}
    \toprule
    Method & \multicolumn{6}{c}{Multi-step Pick and Place} & \multicolumn{3}{c}{Wrap Wire} \\
    
      & \multicolumn{3}{c}{Success Rate \rev{(\%)}} & \multicolumn{3}{c}{Subtasks Completed} & \multicolumn{3}{c}{Success Rate \rev{(\%)}} \\
    \cmidrule(l){2-4}\cmidrule(l){5-7}\cmidrule(l){8-10}
     & 25 & 50 & 100 & 25 & 50 & 100 & 25 & 50 & 100 \\
    \midrule 
    
    No Pretrain ($\targetD$) & 40 & 20 & 30 & 1.15 & 0.9 & 1.15 & 25 & 45 & 35  \\
     No Pretrain ($\priorD + \targetD$) &  45 & 30 & 50 & 1.25 & 1.2 & 1.4 & 15 & 30 & 30 \\
    \midrule
    all-stages & \textbf{55} & \textbf{80} & \textbf{95} & \textbf{1.2} & \textbf{1.8} & \textbf{1.95} & \textbf{25} & \textbf{50} & \textbf{55}\\
    \rev{half-stages} & 45 & 60 & 65 & 1.15 & 1.45 & 1.55 & 5 & 35 & 25\\
     \rev{2-stages} & 35 & 45 & 75 & 1.05 & 1.3 & 1.6 & 20 & \textbf{50} & 40\\
     \rev{1-stage} & \textbf{55} & 65 & 80 & 1.3 & 1.55 & 1.75 & 15 & 15 & 45\\
     \rev{1 stage per domain} & 10 & 50 & 50 & 0.65 & 1.3 & 1.25 & 15 & 15 & 20 \\
    \bottomrule
  \end{tabular}
  % \end{small}
  % }
  \label{tab:real-stages}
\end{table*}

\begin{table*}[htbp]
  \centering
  \caption{\rev{\texttt{sim2real}: Language annotations and language granularity on 2-step real-world pick-and-place}}
  \setlength\aboverulesep{0ex}
  \setlength\belowrulesep{0ex}
  \begin{tabularx}{\textwidth}{|X|X|X|X|}
    \toprule
    \textbf{All-stages} & \rev{\textbf{Half-stages}} & \rev{\textbf{2-stage}} & \rev{\textbf{1-stage}}\\
    \cmidrule(){1-4}
    gripper open, reaching for carrot, out of bowl & \multirow{2}{=}{gripper open, reaching for carrot, out of bowl} & \multirow{7}{=}{picking carrot and putting in bowl} & \multirow{14}{=}{random language embedding} \\
    \cmidrule(){1-1}
    gripper open, moving down over carrot, out of bowl & & & \\
    \cmidrule(){1-2}
    gripper closing, with carrot, out of bowl & gripper closing, with carrot, out of bowl & & \\
    \cmidrule(){1-2}
    gripper closed, moving up with carrot, out of bowl & \multirow{3}{=}{gripper closed, moving up with carrot} & & \\
    \cmidrule(){1-1}
    gripper closed, moving sideways with carrot, out of bowl & & & \\
    \cmidrule(){1-1}
    gripper closed, with carrot, above bowl & & & \\
    \cmidrule(){1-2}
    gripper open, dropped carrot, in bowl & gripper open, dropped carrot, in bowl& & \\
    \cmidrule(){1-3}
    gripper open, reaching for bowl, out of plate & \multirow{2}{=}{gripper open, reaching for bowl, out of plate} & \multirow{7}{=}{picking bowl and putting in plate} & \\
    \cmidrule(){1-1}
    gripper open, moving down over bowl, out of plate & & & \\
    \cmidrule(){1-2}
    gripper closing, with bowl, out of bowl & gripper closing, with bowl, out of plate & & \\
    \cmidrule(){1-2}
    gripper closed, moving up with bowl, out of plate & \multirow{3}{=}{gripper closed, moving up with bowl} & & \\
    \cmidrule(){1-1}
    gripper closed, moving sideways with carrot, out of bowl & & & \\
    \cmidrule(){1-1}
    gripper closed, with bowl, above plate & & & \\
    \cmidrule(){1-2}
    gripper open, dropped bowl, in plate & gripper open, dropped bowl, in plate & & \\
    \bottomrule
    \end{tabularx}
    \label{tab:language-2pp}
\end{table*}

\begin{table*}[htbp]
  \centering
  \caption{\rev{\texttt{sim2real}: Language annotations and language granularity on wire wrap}}
  \setlength\aboverulesep{0ex}
  \setlength\belowrulesep{0ex}
  \begin{tabularx}{\textwidth}{|X|X|X|X|}
    \toprule
    \textbf{All-stages} & \rev{\textbf{half-stages}} & \rev{\textbf{2-stage}} & \rev{\textbf{1-stage}}\\
    \cmidrule(){1-4}
    gripper open, reaching for plug & \multirow{2}{=}{gripper open, reaching for plug} & \multirow{12}{=}{picking and wrapping beads around cylinder} & \multirow{14}{=}{random language embedding} \\
    \cmidrule(){1-1}
    gripper open, moving down over plug & & & \\
    \cmidrule(){1-2}
    gripper closing around plug & \multirow{2}{=}{gripper closing and lifting plug} & & \\
    \cmidrule(){1-1}
    gripper closed, moving up with plug & & & \\
    \cmidrule(){1-2}
    counter-clockwise left & \multirow{4}{=}{counter-clockwise} & & \\
    \cmidrule(){1-1}
    counter-clockwise front & & & \\
    \cmidrule(){1-1}
    counter-clockwise right & & & \\
    \cmidrule(){1-1}
    counter-clockwise back & & & \\
    \cmidrule(){1-2}

    clockwise left & \multirow{4}{=}{clockwise} & & \\
    \cmidrule(){1-1}
    clockwise front & & & \\
    \cmidrule(){1-1}
    clockwise right & & & \\
    \cmidrule(){1-1}
    clockwise back & & & \\
    \cmidrule(){1-3}
    gripper open, above plug with wire fully wrapped & gripper open, above blender with wire fully wrapped & \multirow{2}{=}{beads fully wrapped} & \\
    \cmidrule(){1-2}
    gripper open, above plug with wire fully unwrapped & gripper open, above blender with wire fully unwrapped & & \\
    \bottomrule

    \end{tabularx}
    \label{tab:language-ww}
\end{table*}

\begin{table*}[h]
  \centering
  \caption{\texttt{sim2real}: Finetuning R3M with our method}
  \resizebox{\textwidth}{!}{
  \begin{tabular}{ccccccccccccc}
    \toprule
    Pretraining & \multicolumn{3}{c}{Action-labeled Data} & \multicolumn{6}{c}{Multi-step Pick and Place} & \multicolumn{3}{c}{Wrap Wire}\\
    \cmidrule(r{4pt}){2-4} \cmidrule(l){5-10} \cmidrule(l){11-13}
    & Sim & \multicolumn{2}{c}{Real} & \multicolumn{3}{c}{Success Rate \rev{(\%)}} & \multicolumn{3}{c}{Subtasks Completed} & \multicolumn{3}{c}{Success Rate \rev{(\%)}} \\
    \cmidrule(l){5-7}\cmidrule(l){8-10} \cmidrule(l){11-13}
    &  $\priorD$ &  $\targetDtargetT$ &  $\targetDpriorT$ & 25 & 50 & 100 & 25 & 50 & 100 & 25 & 50 & 100 \\
    \midrule
    R3M + adapters, Lang Reg. & \ding{51} & \ding{51} & -- & 0 & 30 & 40 & 0.7 & 0.95 & 1.25 & 0 & 5 & 5\\
    \midrule 
    Lang Reg. (ours) & \ding{51} & \ding{51} & -- & 55 & \textbf{80} & \textbf{95} & 1.2 & \textbf{1.8} & \textbf{1.95} & 25 & \textbf{50} & \textbf{55}\\
    R3M (frozen) & \ding{51} & \ding{51} & \ding{51} & \textbf{75} & 75 & 85 & \textbf{1.6} & 1.55 & 1.75 & \textbf{30} & 25 & 20\\
    \bottomrule
  \end{tabular}
  }
  \label{tab:r3m-ours-pretrain}
\end{table*}

% \newpage
% \newpage

% \pagebreak
% \afterpage{\clearpage}
% \raggedbottom
\subsection{Combining R3M with Our Approach}
\label{appendix:r3m-ft-ours}
We implemented and evaluated multiple ways to combine R3M with our image-language pretraining to see if it would be possible to leverage the benefits of both R3M's large-scale pretraining and our method's domain-invariant representation learning. Instead of initializing a ResNet from scratch before image-language pretraining, we experiment with using R3M weights and finetuning the last layer, the entire network, or inserted convolutional adapter modules~\cite{chen2024convadapter}. Finetuning adapters (denoted R3M$+$adapters) performs the best in \texttt{sim2sim}, matching the performance of our method on 2-step pick-and-place.

Based on this sim2sim performance, we evaluated R3M$+$adapters on \texttt{sim2real}, but this performed worse than either frozen R3M or our method in \texttt{sim2real} (Table~\ref{tab:r3m-ours-pretrain}). We hypothesize that this is because during image-language pretraining on both sim and real images, the trainable adapters learn to pick out features primarily in the simulation images as this is out-of-distribution for R3M which was trained on real-world videos, which enables R3M$+$adapters to do well in \texttt{sim2sim} but not \texttt{sim2real}.

\end{document}